\documentclass{article}

\usepackage[utf8]{inputenc}
\usepackage{booktabs} 
\usepackage{multirow}
\usepackage{longtable}
\usepackage{graphicx}
\usepackage{subcaption}
\usepackage{bm}
\usepackage{amsmath}
\usepackage{xcolor} 
\usepackage{physics}

\usepackage{xcolor}

\begin{document}

\title{A categorisation and implementation of digital pen features for behaviour characterisation}

\author{Alexander Prange
    \and Michael Barz
    \and Daniel Sonntag
    }

\date{DFKI}

\maketitle

\begin{abstract}
In this paper we provide a categorisation and implementation of digital ink features for behaviour characterisation. Based on four feature sets taken from literature, we provide a categorisation in different classes of syntactic and semantic features. We implemented a publicly available framework to calculate these features and show its deployment in the use case of analysing cognitive assessments performed using a digital pen.
\end{abstract}

\section{Introduction}
The research described in this paper is motivated by the development of applications for the behaviour analysis of handwriting and sketch input.
Our goal is to provide other researchers with a reproducible, categorised set of features that can be used for behaviour characterisation in different scenarios.
We use the term feature to describe properties of strokes and gestures which can be calculated based on the raw sensor input from capture devices, such as digital pens or tablets.

In this paper, a large number of features known from the literature are presented and categorised into different subsets.
For better understanding and reproducibility we formalised all features either using mathematical notations or pseudo code and summarised them in the appendix section of this paper.
Furthermore, we created a open-source python reference implementation of these features, which is publicly available\footnote{Download is available at GitHub https://github.com/DFKI-Interactive-Machine-Learning/ink-features}.

The presented ink features can be used in a variety of ways.
Most commonly they are used to perform character and gesture recognition based on machine learning techniques.
Here we describe their use for automated behaviour characterisation in the use case of cognitive assessments.
Traditionally these tests are performed using pen and paper with manual evaluation by the therapist.
We show how ink features can be used in that context to provide additional feedback about the cognitive state of the patient.
Finally, we explain how digital ink can be used as an input modality in multimodal, multisensor interfaces.

\section{Digital Ink}
Over the past few years the availability of digital pen hardware has increased drastically, and there is a wide variety of devices to choose from if dealing with handwriting analysis.
Several different technologies are used to record handwriting, e.g., accelerometer-based digital pens convert the movement of the pen on the surface whereas active pens transmit their location, pressure and other functionalities to the built-in digitiser of the underlying device.
Positional pens, most often encountered in graphic tablets, have a surface that is sensitive to the pen tip.
A special, nearly invisible, dot pattern can be printed on regular paper, so that camera-based pens detect where the stylus contacts the writing surface.

In this work we focus on the similarities between the most commonly used hardware devices for sketch recognition.
As not all technologies deliver the same type of sensor data, we identified a subset that is covered by the majority of input devices.
We refer to it as digital ink, a set of time-series data containing coordinates and pressure at each timestamp.
For the remainder of this paper we use the follwing notation:

\begin{equation}
    x, y : \text{coordinates}
\end{equation}
\begin{equation}
    p : \text{pressures}
\end{equation}
\begin{equation}
    t: \text{timestamps}
\end{equation}

A series \(S\) of \(n\) sample points between a pen down and pen up event is called a stroke and can be represented as a series of tuples

\begin{equation}
    S = (x_0, y_0, p_0, t_0), (x_1, y_1, p_1, t_1), ..., (x_{n-1}, y_{n-1}, p_{n-1}, t_{n-1})
\end{equation}

where $x_i$ represents the x coordinate of the i-th sample point within the series, with $0 \leq i < n$. The tuple itself may be referenced by \(s_i\). Timestamps are measured in milliseconds, it is insignificant if they are absolute or relative to the first point.

\section{Features}
We refer to individual, measurable properties or charateristics of digital ink as features. Features are calculated directly from the input sample points and represented by a numerical value. Therefore a feature can be seen as a function:

\begin{equation}
    f: S \mapsto {\rm I\!R}
\end{equation}

Depending on the feature, $S$ can be a set of strokes (gesture level), a single stroke (stroke level) or a subset of sample points. Usually a vector of features
\begin{equation}
    F = [f_1, ... , f_{m}]
\end{equation}
is extracted from the input gesture and can then be used in a classifier.

\subsection{Feature Sets}
Traditionally stroke level features are most often used for statistical gesture recognition. One of the most prominent set of features was presented by Dean Rubine in 1991 \cite{RUBINE}. It contains a total of 13 features that have been designed to reflect the visual appearance of strokes in order to be used in a gesture recogniser. 
More recent work by Don J.M. Willems and Ralph Niels \cite{WILLEMS} defines a total of 89 features using formal mathematical descriptions and algorithms.
Adrien Delaye and Eric Anquetil introduced the HBF49 Feature Set \cite{HBF49}, which contains 49 features and was specifically designed for different sets of symbols and as reference for evaluating symbol recognition systems.
In previous work we used 14 features described by Sonntag et al. \cite{WEBER} to distinguish between written text and other types of gestures in online handwriting recognition.

\subsubsection{Common Features}
Due to the nature of sketched or handwritten input there are a few features and concepts that the above mentioned publications have in common. The most prominent example is the length of a stroke, here we use the Euclidean distance to measure the distance between sampling points. 

Given two sampling points $q = (x_q, y_q)$ and $r = (x_r, y_r)$ their distance is calculated as follows:

\begin{equation}
    \| q r \| = \| r - q \| = \sqrt{(x_r - x_q)^2 + (y_r - y_q)^2}\end{equation}

The length of a stroke (a squence of sampling points) is given by the sum of distances between the sampling points:

\begin{equation}f_{Stroke Length} = \sum_{i=1}^{n-1} ||s_{i} - s_{i-1} ||\end{equation}

\begin{figure}
    \centering
    \includegraphics[width=0.6\linewidth]{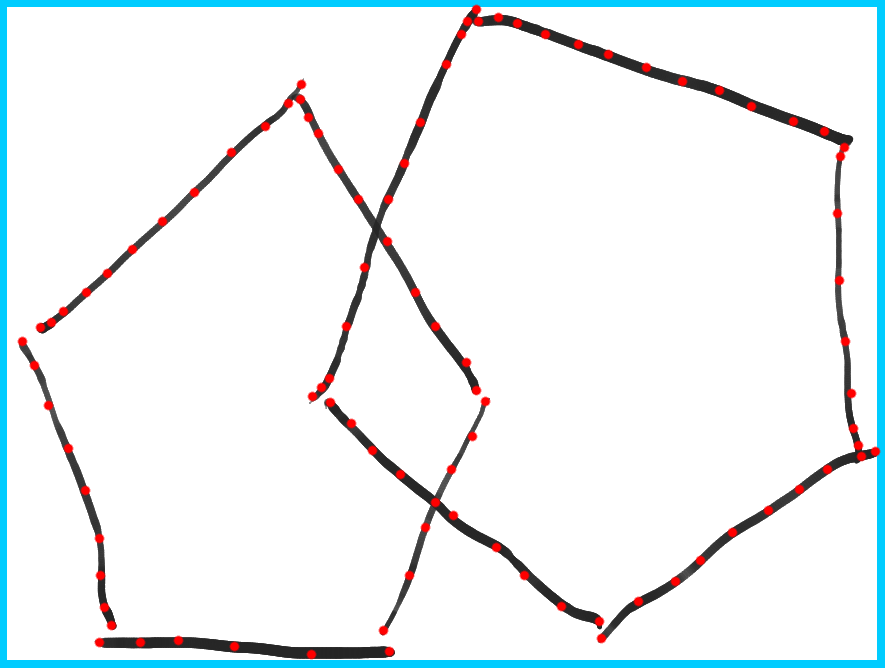}
    \caption{The rectangular bounding box (cyan) around a set of strokes (black) given by a set of sample points (red).}
    \label{fig:boundingbox}
\end{figure}

A bounding box (see figure \ref{fig:boundingbox}) around a set of strokes describes the smallest enclosing area containing the entire set of points. Its size is determined by the minimum and maximum sample points:

\begin{equation}x_{min} = \min_{0 \leq i < n} x_i\end{equation}
\begin{equation}y_{min} = \min_{0 \leq i < n} y_i\end{equation}
\begin{equation}x_{max} = \max_{0 \leq i < n} x_i\end{equation}
\begin{equation}y_{max} = \max_{0 \leq i < n} y_i\end{equation}

The area of the bounding box is then given by:

\begin{equation}f_{Bounding Box Area} = (x_{max} - x_{min}) \cdot (y_{max} - y_{min})\end{equation}

\subsection{Feature Categories}


We have chosen the above described sets of features which are formalised in a reproducable way.
As the features describe different aspects of the digital ink we decided to sort them into different categories. 
We distinguish each feature to be either a syntactic or semantic feature. 
Syntactic features reflect task independent characteristics about the geometry of the input, whereas semantic features describe closely task related knowledge.
In this work we introduce 7 categories of syntactic features:

\subsubsection{Angle Based}
Angle based features are calculated from angles between sample points (e.g., curvature, perpendicularity, rectangularity).

\begin{equation}
    \theta _ { i } = \arccos { \bigg\{ \dfrac{\overrightarrow{s_{i-1}s_{i}} \cdot \overrightarrow{s_{i}s_{i+1}}}     {||\overrightarrow{s_{i-1}s_{i}}|| \cdot ||\overrightarrow{s_{i}s_{i+1}}||} \bigg\} } 
\end{equation}

\begin{equation}
    f _ { Curvature } = \sum _ { i = 1 } ^ { n - 2 } \theta _ { i } 
\end{equation}
\begin{equation}
    f _ { Perpendicularity } = \sum _ { i = 1 } ^ { n - 2 } \sin ^ { 2 } \left( \theta _ { i } \right)
\end{equation}

\subsubsection{Space Based}
Space based features depend on the distances between samples (e.g., convex hull area, principal axes, compactness). The area $A$ of a gesture is usually derived from the area of the convex hull around all sample points, which can be calculated using Graham's algorithm~\cite{graham1972}.

\begin{equation}
    f_{ConvexHullArea} = A
\end{equation}

With the area of the convex hull and the length of its perimeter $l$ we get a feature called compactness. The closer the sample points are together, the smaller the compactness will be. Handwritten texts, e.g., will have a larger compactness than geometric symbols, such as rectangles~\cite{WILLEMS}.

\begin{equation}
    f _ { Compactness } = \frac { l ^ { 2 } } { A }
\end{equation}

Related to the bounding box of a figure, we use its side length to calculate the eccentricity. Note that we are using the co-ordinate axes instead of the principal axes (which are rotated with the pen gesture).

\begin{equation}
    f _ { Eccentricity } = \sqrt { 1 - \frac { b ^ { 2 } } { a ^ { 2 } } }
\end{equation}

\subsubsection{Centroidal}
Centroidal features describe relations between sample points and the overall centroid (e.g., centroid offset, deviation, average radius).

Using the dimensions of the bounding box we calculate the center point $c$:

\begin{equation}
    c = \binom{x_{center}}{y_{center}} = \binom{x_{min} + 0.5 \cdot (x_{max} - x_{min})}{y_{min} + 0.5 \cdot (y_{max} - y_{min})}
\end{equation}

The average distance of sample points to the center point is another feature:

\begin{equation}
    f _ { MeanCentroidDistance} = \frac { 1 } { n } \sum _ { i = 0 } ^ { n - 1 } \left\| s _ { i } - \mathbf { c } \right\|
\end{equation}

\subsubsection{Temporal}
Temporal features are derived from timestamps of sample points (e.g., duration, speed, acceleration). The velocity $v$ between sample points is defined as:

\begin{equation}
    v _ { i } = \frac { s _ { i + 1 } - s _ { i - 1 } } { t _ { i + 1 } - t _ { i - 1 } }
\end{equation}

From which the feature of average velocity is calculated:

\begin{equation}
    f _ { AverageVelocity } = \frac { 1 } { n - 2 } \sum _ { i = 1 } ^ { n - 2 } \left\| \mathbf { v } _ { i } \right\|
\end{equation}

The acceleration is calculated as follows:
\begin{equation}
    a _ { i } = \frac { v _ { i + 1 } - v _ { i - 1 } } { t _ { i + 1 } - t _ { i - 1 } } 
\end{equation}

And the average acceleration is then given by:

\begin{equation}
    f _ { AverageAcceleration } = \frac { 1 } { n - 4 } \sum _ { i = 2 } ^ { n - 3 } \left\| \mathbf { a } _ { i } \right\|
\end{equation}

\subsubsection{Pressure Based}
Pressure based features are computed from hardware sensors capturing applied pressure (e.g., average pressure, standard deviation). The most intuitive and obvious features are the average pressure and the standard deviation in pressure:

\begin{equation}
    f _ { AveragePressure } = \frac { 1 } { n } \sum _ { i = 0 } ^ { n - 1 } p _ { i } 
\end{equation}

\begin{equation}
    f _ { StandardPressureDeviation } = \sqrt { \frac { 1 } { n } \sum _ { i = 0 } ^ { n - 1 } \left( p _ { i } - f _ { AveragePressure } \right) ^ { 2 } } 
\end{equation}

\subsubsection{Trajectory Based}
Trajectory based features reflect the visual appearance of strokes (e.g., closure, average stroke direction).

The path length from one sample point to another is denoted $L$ and is calculated as follows:
\begin{equation}
    L _ { i , j } = 
        \sum _ { k = i } ^ { j - 1 } 
            \left\{ 
                \begin{array} { l }
                    { 0 } \\ 
                    { \left\| s _ { k } s _ { k + 1 } \right\| } 
                \end{array} 
            \right.
\end{equation}

$L = L _ { 0 , n - 1 }$ is the total length of $S$. Whereas the first to last point vector and its length is: \begin{equation}v = \overrightarrow { s _ { 1 } s _ { n } } , \quad \| v \| = \left\| s _ { 1 } s _ { n } \right\|\end{equation}

Typical trajectory based features are closure and average direction:

\begin{equation}
    f _ { Closure } = \frac { \| v \| } { L }
\end{equation}

\begin{equation}
    f _ { AverageDirection } = \frac { 1 } { n - 1 } \sum _ { i = 0 } ^ { n - 2 } \arctan \left( \frac { y _ { i + 1 } - y _ { i } } { x _ { i + 1 } - x _ { i } } \right)
\end{equation}

\subsubsection{Meta}
Meta features are higher level features and relations between components (e.g., number of strokes, inter-connections, crossings, straight line ratio). One intuitive example would be the number of straight lines ($f_{\#StraightLines}$) or to be more precise the number of straight segments. We use a sliding window with a threshold to calculate sets of connected points which have minimal curvature between them. The size sliding window and threshold can be either dynamically adjusted to the length of the stroke or be a fixed value depending on the task.

The feature called \emph{connected components} ($f_{ConnectedComponents}$) ~\cite{WILLEMS} describes the number of segments which are interconnected with other segments, e.g., crossings between strokes.

\begin{center}
\begin{longtable}{|c|c|c|}

\hline
\multicolumn{2}{|l|}{Geometric Features}\\
\hline
Angle Based                             & Space Based                             \\
\hline
Circular Variance                       & Stroke Length                           \\ 
Rectangularity                          & Gesture Length                          \\ 
Curvature                               & Perimeter Length                        \\ 
Average Curvature                       & Compactness                            \\ 
SD of Curvature                         & Eccentricity                           \\ 
Angles after Resampling                 & Principal Axes                          \\ 
Cosine of First to Last Point Vector    & First Point X                           \\ 
Sine of First to Last Point Vector      & First Point Y                          \\ 
Cosine Initial Vector                   & Last Point X                            \\ 
Sine Initial Vector                     & Last Point Y                            \\ 
Bounding Box Diagonale Angle            & First to Last Point Vector              \\ 
Perpendicularity                        & 2D Histogramm         \\ 
Average Perpendicularity                & Ratio of Axes                           \\ 
SD of Perpendicularity                  & Ratio of Principal Axes                 \\ 
Signed Perpendicularity                 & Length of First Principal Axis          \\ 
K-Perpendicularity                      & SD of Stroke Length                     \\ 
Maximum k-Angle                         & Sample Ratio Octants                    \\ 
Absolute Directional Angle              & Convex Hull Area                        \\ 
Relative Angle Histogram                & Convex Hull Compactness                 \\ 
Principal Axis Orientation (sin)        & Distance of First to Last Point         \\ 
Principal Axis Orientation (cos)        & Average Length of Straight Lines        \\ 
Maximum Angular Difference              & Initial Horizontal Offset               \\
Circular Variance                       & Final Horizontal Offset                 \\ 
Sum of Absolute Values of Angles        & Initial Vertical Offset                 \\ 
Sum of Angles                           & Final Vertical Offset                   \\ 
Sum of Squared Angles                   &                             \\ 
Macro Perpendicularity                  &                                \\ 
Average Macro Perpendicularity          &                         \\ 
SD of Macro Perpendicularity            &               \\ 
Absolute Curvature                      &                 \\ 
Squared Curvature                       &                              \\
\hline
\multicolumn{2}{|c|}{Centroidal}\\
\hline
\multicolumn{2}{|c|}{Deviation}\\
\multicolumn{2}{|c|}{Centroid Offset}\\
\multicolumn{2}{|c|}{Average Centroidal Radius}\\
\multicolumn{2}{|c|}{SD of Centroidal Radius}\\
\multicolumn{2}{|c|}{Hu moments}\\
\hline\hline

\multicolumn{2}{|l|}{Temporal Features}\\
\hline
\multicolumn{2}{|c|}{Maximum Speed (Squared)} \\
\multicolumn{2}{|c|}{Duration of Gesture} \\
\multicolumn{2}{|c|}{Pen Up/Pen Down Ratio} \\
\multicolumn{2}{|c|}{Average Velocity} \\
\multicolumn{2}{|c|}{SD of Velocity} \\
\multicolumn{2}{|c|}{Maximum Velocity} \\
\multicolumn{2}{|c|}{Average Acceleration} \\
\multicolumn{2}{|c|}{SD of Acceleration} \\
\multicolumn{2}{|c|}{Maximum Acceleration} \\
\multicolumn{2}{|c|}{Maximum Deceleration} \\
\hline\hline

\multicolumn{2}{|l|}{Pressure Based Features}\\
\hline
\multicolumn{2}{|c|}{Average Pressure} \\
\multicolumn{2}{|c|}{SD of Pressure} \\
\hline\hline

\multicolumn{2}{|l|}{Trajectory Features}\\
\hline
\multicolumn{2}{|c|}{Closure} \\
\multicolumn{2}{|c|}{Inflexion X} \\
\multicolumn{2}{|c|}{Inflexion Y} \\
\multicolumn{2}{|c|}{Proportion of Downstroke Trajectory} \\
\multicolumn{2}{|c|}{Ratio between Half-Perimeter and Trajectory} \\
\multicolumn{2}{|c|}{Average Stroke Direction} \\
\multicolumn{2}{|c|}{Cup Count} \\
\multicolumn{2}{|c|}{Last Cup Offset} \\
\multicolumn{2}{|c|}{First Cup Offset} \\
\multicolumn{2}{|c|}{Number of Pen Down Events} \\
\multicolumn{2}{|c|}{Sin Chain Code} \\
\multicolumn{2}{|c|}{Cos Chain Code} \\
\multicolumn{2}{|c|}{SD of Stroke Direction} \\
\hline\hline

\multicolumn{2}{|l|}{Meta Features}\\
\hline
\multicolumn{2}{|c|}{Number of Strokes}\\
\multicolumn{2}{|c|}{Number of Straight Lines}\\
\multicolumn{2}{|c|}{SD of Straight Lines}\\
\multicolumn{2}{|c|}{Straight Line Ratio}\\
\multicolumn{2}{|c|}{Largest Straight Line Ratio}\\
\multicolumn{2}{|c|}{Number of Connected Components}\\
\multicolumn{2}{|c|}{Number of Crossings}\\
\hline

\caption{Categorisation of syntactic features into classes.} \label{tab:features_summary} \\

\end{longtable}
\end{center}

\subsection{Semantic/Task Based Features}
Depending on the task, additional features can be deduced from the task itself. As these features describe higher level semantic concepts about the sketched contents, we often refer to them as semantic features. Semantic features highly depend on the given context and therefore vary noticeably between different tasks. Such features usually cannot be transfered easily to other tasks, as they are often hard-coded per task.

\begin{figure}
    \centering
    \includegraphics[width=.7\linewidth,trim={8.1cm 1.2cm 8.5cm 1.2cm},clip]{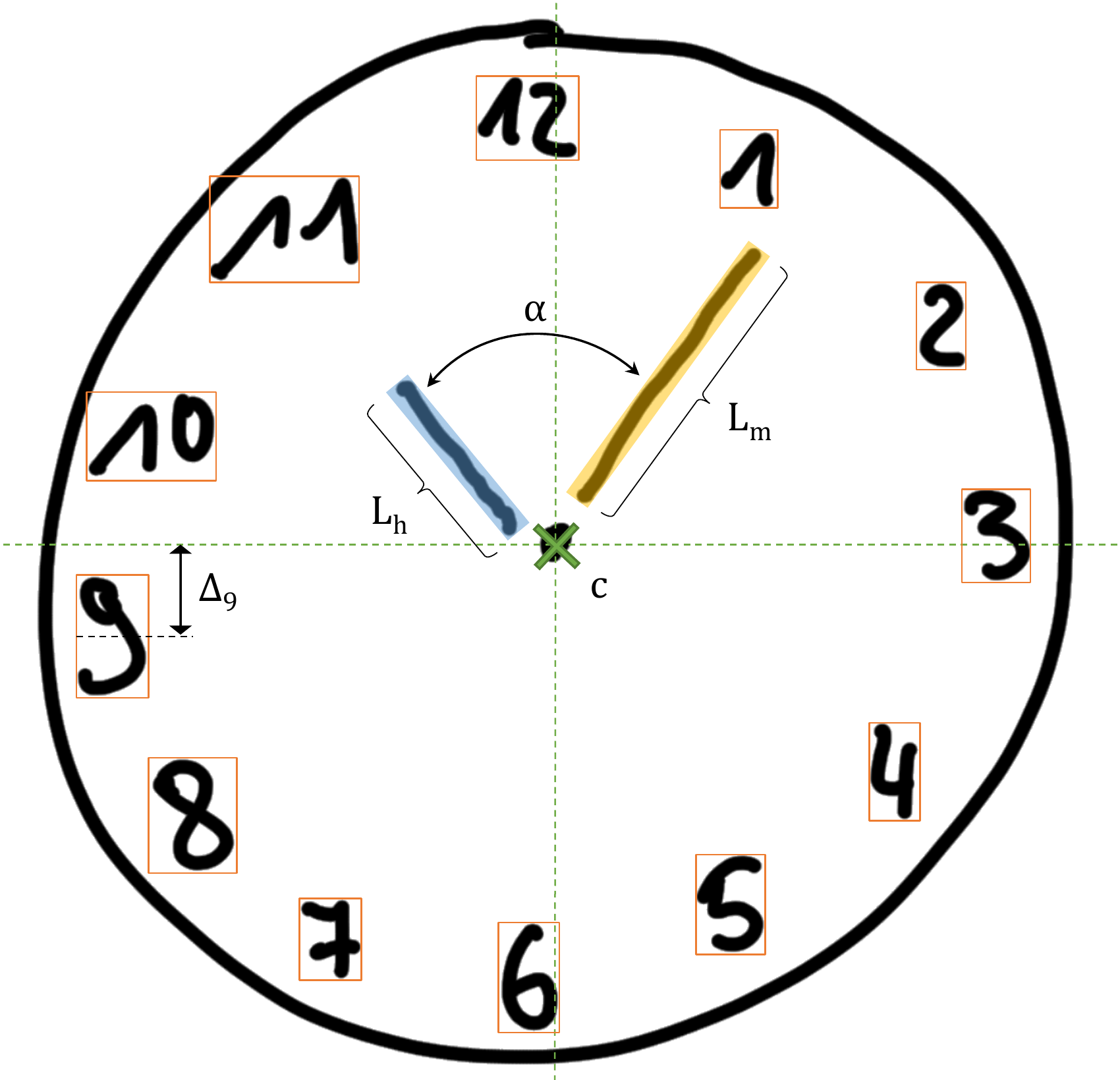}
    \caption{Visualisation of semantic features in the context of the Clock Drawing Test.}
    \label{fig:semantic_features}
\end{figure}

Figure \ref{fig:semantic_features} shows the visualisation of a selected semantic feature set in the context of the Clock Drawing Test, a widely used pen and paper screening test used for more than 50 years as a screening tool for cognitive impairment. Participants are asked to draw a clock face with the time set to 10 past 11 o'clock. The drawn clock is then examined by a trained physician and rated based on a predefined scoring scheme, reflecting the visual appearance and integrity of the clock using a numerical score. In this example we deduced the following features based on the traditional scoring system:
\begin{itemize}
  \item $c$ denotes the center point of the clock (centroid), the closer it is to the center of the clock's circle, the more points are awarded.
  \item $L_h$ and $L_m$ represent the lengths of the hour and minute hands respectively. If the clock is well drawn, the hour hand should be shorter than the minute hand.
  \item The angle between the hour and minute hands is denoted as $\alpha$, together with the orientation of the hands it can be used to determine if the correct time was set.
  \item $\Delta_9$ is the displacement of clock face digits relative to their ideal location. In this example it is the vertical offset of digit number 9 to its correct center position.
\end{itemize}

\section{Related Work}
One of the first reproducible ink feature sets was presented by Rubine~\cite{RUBINE} in 1991. He described how to use these features in a trainable single-stroke recogniser for gestures.
Willems and Niels~\cite{WILLEMS} presented a set of 89 ink features which they used for iconic and multi-stroke gesture recognition.
The HBF49 feature set was presented by Delaye and Anquetil~\cite{HBF49} to be used in online symbol recognition.
Sonntag et al.~\cite{WEBER} used ink features to distinguish between writing and sketching in online mode detection of handwriting input.

Ink features can not only be used for gesture or sketch recognition, but also for characterisation of handwriting behaviour.
Drotar et al.~\cite{parkinsons_in_air} have shown that the analysis of in-air movement can be used as a marker for Parkinson's disease.
The kinematic analysis of handwriting movements can be used to distinguish between certain forms of dementia \cite{kinematic_analysis}.

Digitalising popular existing cognitive assessments, such as the Clock Drawing Test (CDT), has been topic of recent debate.
There are clear benefits resulting from digitalisation, such as increased diagnostic accuracy \cite{increased_diagnostic_accuracy}.
Davis et al. only recently presented their work on how to infer congitive status from subtle behaviours observed in digital ink \cite{think_inferring}.
Based on such ink features machine learning models can be trained \cite{learning_classification_models}, which can also be explained by existing, validated scoring schemes \cite{interpretable_ml_models}.
Examples of complex digitalised cognitive assessments include the Rey-Osterrieth Complex Figure test \cite{automated_rocf}, which can be used for various purposes, such as diagnosing the periphery \cite{diagnosing_the_periphery}.

Behaviour characterisation can be also used in different settings, e.g., to gain feedback about cognitive load of the writer.
Luria and Rosenblum \cite{computerized_multidimensional} conducted a study to determine the effect of mental workload on handwriting behaviour.
Yu et al.~\cite{mental_workload_classification} showed that online writing features can be used for mental workload classification, such as congitive load evaluation~\cite{cognitive_load_evaluation}.
Ink features can be also used in multimodal scenarios~\cite{multi_modal_handbook}, where they may enhance the prediction of cognitive and emotional states~\cite{cognitive_load_measurement}.

\section{Use Case: Cognitive Assessments}

On use case where we apply our feature set is the analysis of handwriting behaviour for dementia screening tools in the Interakt project \cite{DBLP:journals/corr/abs-1709-01796}.
Dementia is a general term for a decline in mental ability severe enough to interfere with daily life.
In 2018, the Alzheimer's Association documented that approximately 10-20\% of the population over 65 years of age suffer from some form of dementia \cite{alzheimers}.
Screening tests for dementia have been the subject of recent debate because there are limitations when they are conducted using pen and paper. 
For example, the collected material is monomodal (written form) and there is no direct digitalisation for further and automatic processing, the results can be biased.
We selected the assessments based on feedback from domain experts and a recent market analysis of existing, most widely used, cognitive assessments conducted by Niemann et al.~\cite{niemann}.
Our selected and implemented paper and penicl tests are shown in table \ref{table:assessments}, namely Age-Concentration (AKT)~\cite{AKT}, Clock Drawing Test (CDT)~\cite{CDT}, CERAD Neuropsychological Battery~\cite{CERAD}, Dementia Detection (DemTect)~\cite{DemTect}, Mini-Mental State Examination (MMSE)~\cite{MMSE}, Montreal Cognitive Assessment (MoCA)~\cite{MoCA}, Rey-Osterrieth Complex Figure (ROCF)~\cite{automated_rocf}, and Trail Making Test (TMT)~\cite{TMT}. 
The selection of the tests accounts for a variety of patient populations and test contexts.

\begin{table*}
  \begin{tabular}{ l c c c c c }
    name & time needed & pen input & symbols\\
    \hline 
    \\
    AKT~\cite{AKT} & 15 min &  100\% & cross-out\\   
    CDT~\cite{CDT} & 2-5 min &  100\% & clock, digits, lines \\
    CERAD~\cite{CERAD} & 30-45 min &  20\% & (see figure \ref{fig:cerad_symbols})\\
    DemTect~\cite{DemTect} & 6-8 min &  20\% & numbers, words \\
    MMSE~\cite{MMSE} & 5-10 min &  9\% & pentagrams \\
    MoCA~\cite{MoCA} & 10 min &  17\% & clock, digits, lines \\
    ROCF~\cite{automated_rocf} & 15 min & 100\% & circles, rectangles, triangles, lines \\
    TMT~\cite{TMT} & 3-5 min & 100\% & lines \\
  \end{tabular}
  \caption{Comparison of the most widely used cognitive assessments}
  \label{table:assessments}
\end{table*}

One of the most prominent example is the internationally used Mini-Mental State Examination (MMSE) \cite{MMSE}, a 30-point questionnaire, which is extensively used in medicine and research to measure cognitive impairment.
Depending on the experience of the physician and the cogntive state of the patient the administration of the test takes between 5 and 10 minutes and examines functions including awareness, attention, recall, language, ability to follow simple commands and orientation \cite{modernMMSE}.
Due to its standardisation, validity, short administration period and ease of use, it is widely used as a reliable screening tool for dementia \cite{mmse_screeningtool}.
The MMSE also includes several tasks which involve handwriting input by the participant, e.g., writing a complete sentence and copying a geometric figure.

\begin{figure}
    \centering
    \includegraphics[width=0.6\linewidth]{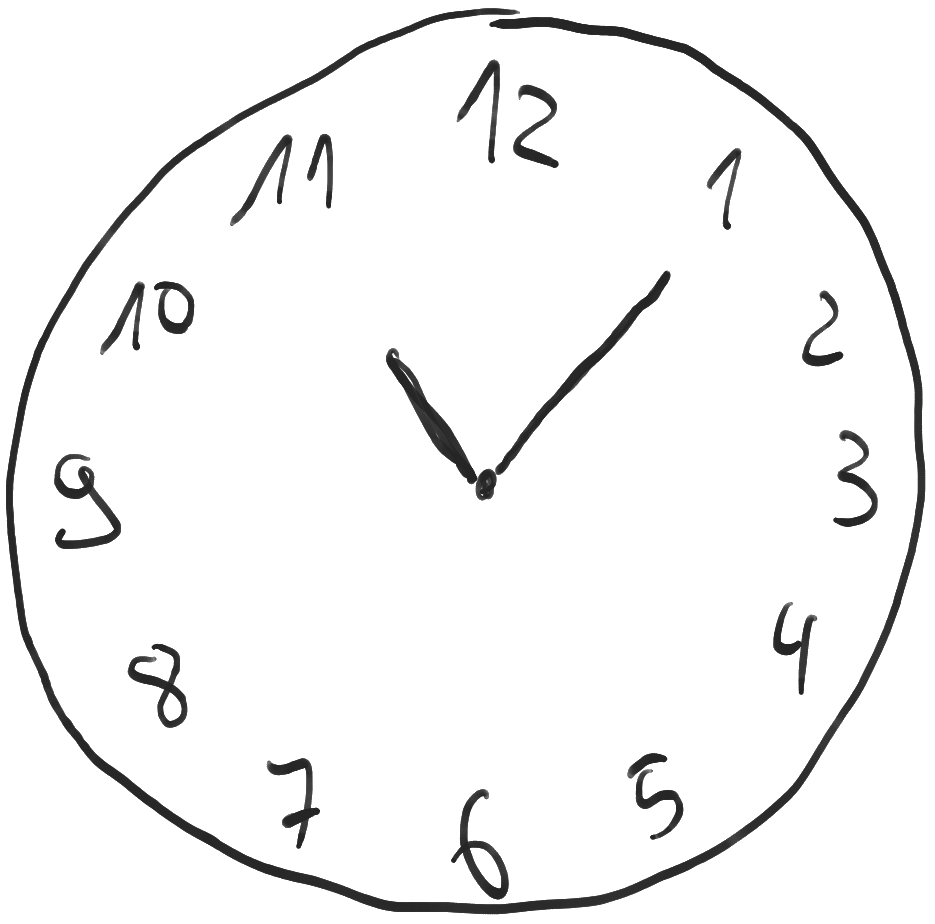}
    \caption{Clock Drawing Test (CDT).}
    \label{fig:clockdrawingtest}
\end{figure}

\begin{figure}
    \centering
    \includegraphics[width=0.4\linewidth]{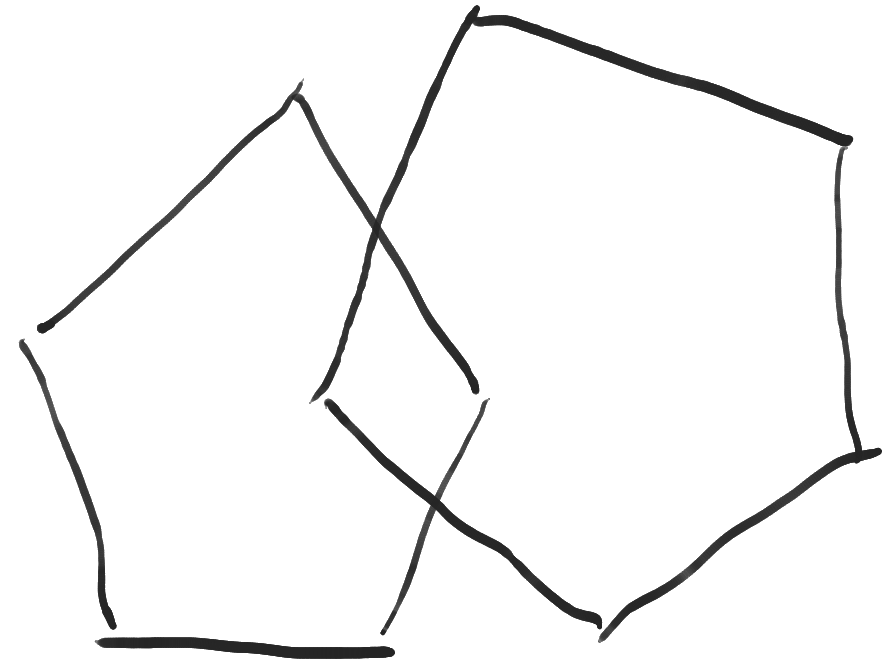}
    \caption{Mini Mental Status Exam (MMSE): Copy pentagram figure task.}
    \label{fig:mmse_pentagrams}
\end{figure}

\begin{figure}
  \centering  
  \begin{subfigure}[b]{0.19\linewidth}
    \includegraphics[width=\linewidth]{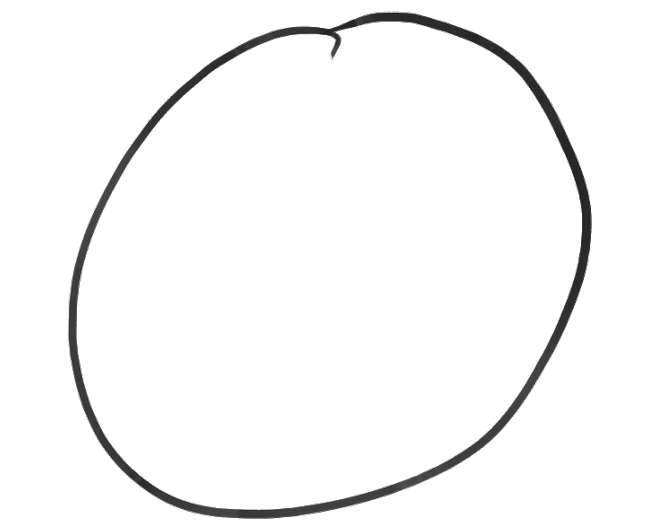}
     \caption{Circle}
  \end{subfigure}
  \begin{subfigure}[b]{0.19\linewidth}
    \includegraphics[width=\linewidth]{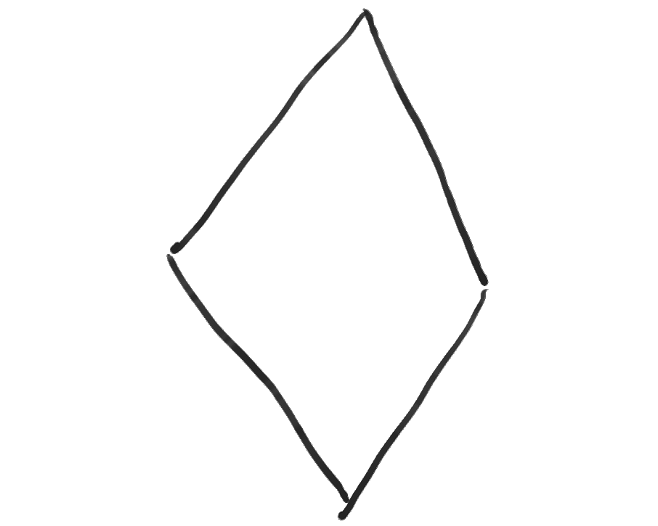}
    \caption{Diamond}
  \end{subfigure}
  \begin{subfigure}[b]{0.19\linewidth}
    \includegraphics[width=\linewidth]{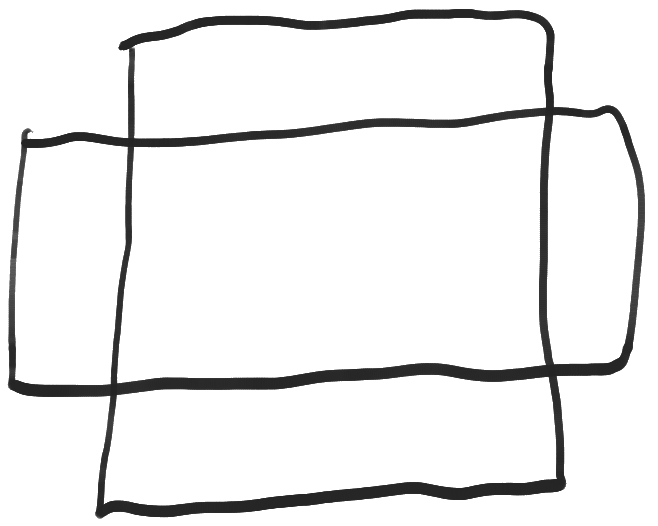}
    \caption{Rectangles}
  \end{subfigure}
  \begin{subfigure}[b]{0.19\linewidth}
    \includegraphics[width=\linewidth]{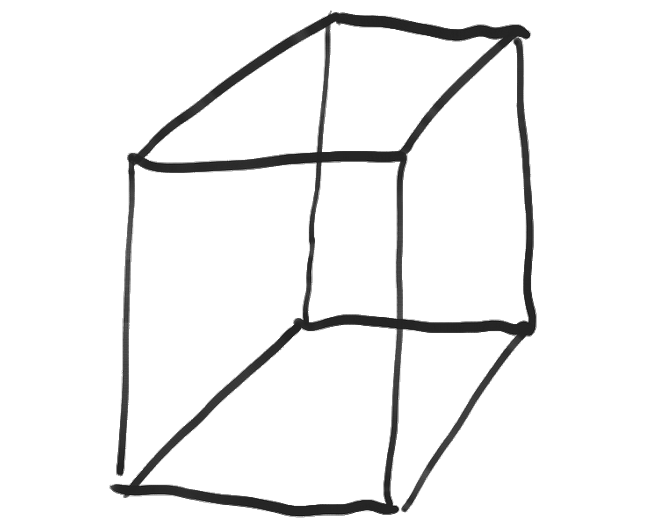}
    \caption{Cube}
  \end{subfigure} 
  \begin{subfigure}[b]{0.19\linewidth}
    \includegraphics[width=\linewidth]{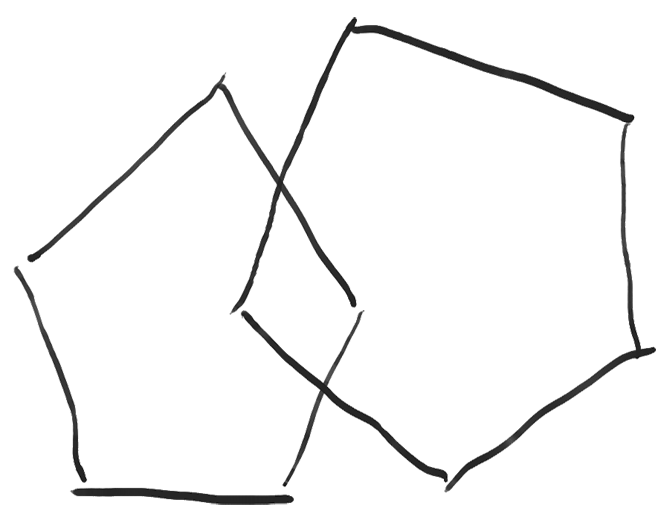}
    \caption{Pentagrams}
  \end{subfigure}  
  \caption{Symbols used in the CERAD neuropsychological battery.}
  \label{fig:cerad_symbols}
\end{figure}

The Clock Drawing Test (CDT) \cite{CDT} is another popular cognitive assessment, where the patient is asked to draw a clock with a specified time on a piece of paper, see figure \ref{fig:clockdrawingtest}.
Based on the completeness and appearance of the clockface and the arrangement of the digits a score is calculated.
The CDT and MMSE are perfect examples for illustrating the two categories of traditional paper and pencil cognitive testing.
There are assessments, like the CDT, which rely solely on handwriting and sketch input to produce a score, whereas there are others, such as the MMSE, which also include other modalities, such as speech for instance.
Depending on the assessment, the handwriting input has a different weight for the overall scoring of the test.
Table \ref{table:assessments} shows the absolute percentages of the test questions where the a pen is used to answer them. 
Tasks in the MMSE containing pen input include writing a sentence and copying a figure of two overalapping pentagrams (see figure \ref{fig:mmse_pentagrams}).
Out of 22 possible points in the scoring of the MMSE, the pen input related task add up to 2 points, resulting in an overall 9\% of the entire test to be scored through analysis of pen input.
Regarding task design the Montreal Cognitive Assessment (MoCA) \cite{MoCA} is comparable with the MMSE and CDT, e.g., it also includes copying a figure and drawing a clock.
The CERAD Neuropsychological Battery \cite{CERAD} is a collection of several tests (including the MMSE and TMT), where amongst others the subject has to copy several shapes depicted in figure \ref{fig:cerad_symbols}.
In the Trail Making Test (TMT) \cite{TMT} the subject has to connect numbers and letters in ascending order.
A more complex example of a test that is rated entirely based on pen input is the Rey-Osterrieth complex figure test (ROCF) \cite{automated_rocf}, where subjects are required to copy the figure three times, once while looking at the template, once directly after that, but without seeing the template, and once from recall 30 minutes later.
The Age-Concentration Test (AKT) \cite{AKT} asks subjects to cross out a specific shape from a set of similar, yet varying shapes in a limited amount of time.
Handwritten words and digits are contained in the DemTect \cite{DemTect}, where subjects translate numbers into words and vice versa.

\begin{figure}
    \centering
    \includegraphics[width=0.5\linewidth]{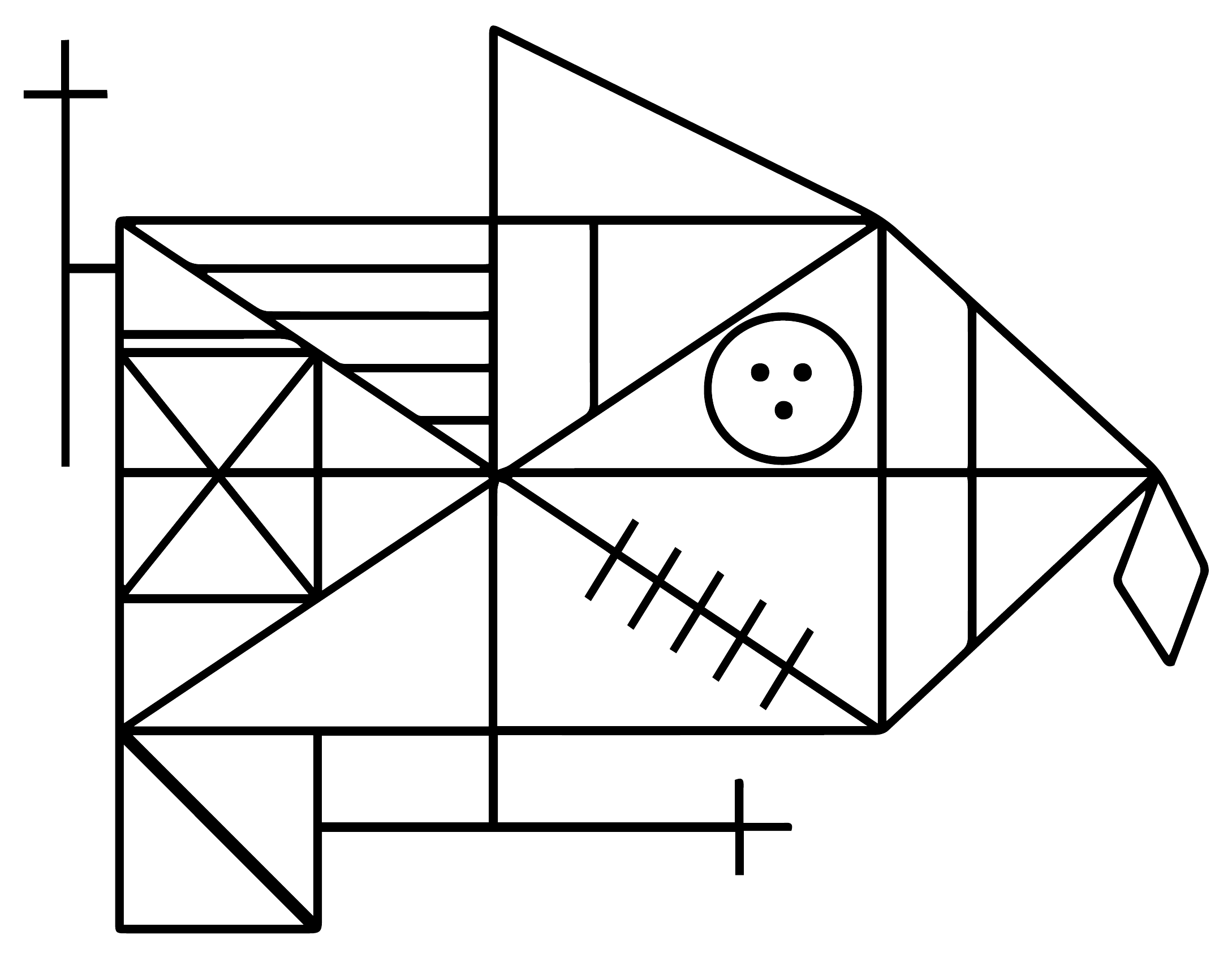}
    \caption{The Rey-Osterrieth complex figure (ROCF).}
    \label{fig:rocf}
\end{figure}

\subsection{Symbols data set}
\begin{figure}
    \centering
    \includegraphics[width=0.8\linewidth,trim={2.8cm 2.1cm 4.2cm 1.2cm},clip]{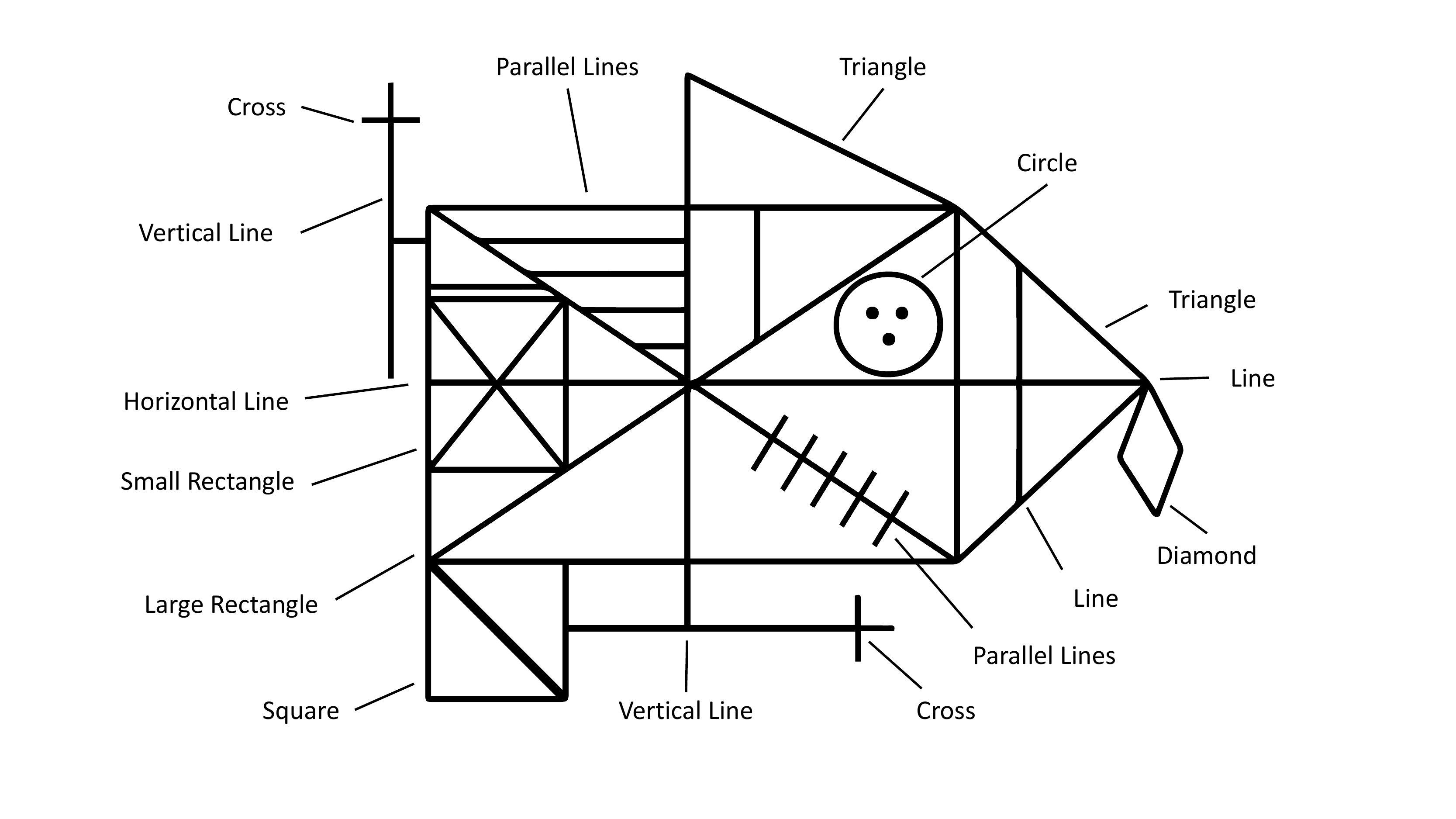}
    \caption{The Rey-Osterrieth complex figure (ROCF) is composed of several sub-shapes.}
    \label{fig:rocf_annotated}
\end{figure}

\begin{figure}
  \centering
  \begin{subfigure}[b]{0.2\linewidth}
    \includegraphics[width=\linewidth]{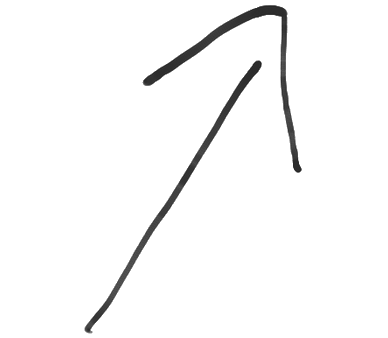}
     \caption{Arrow}
  \end{subfigure}
  \begin{subfigure}[b]{0.2\linewidth}
    \includegraphics[width=\linewidth]{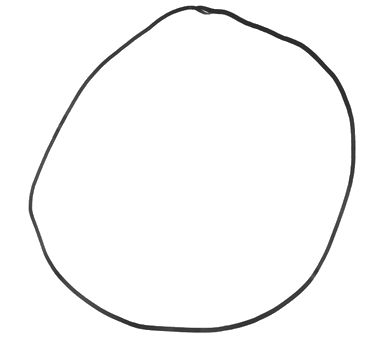}
    \caption{Circle}
  \end{subfigure}
  \begin{subfigure}[b]{0.2\linewidth}
    \includegraphics[width=\linewidth]{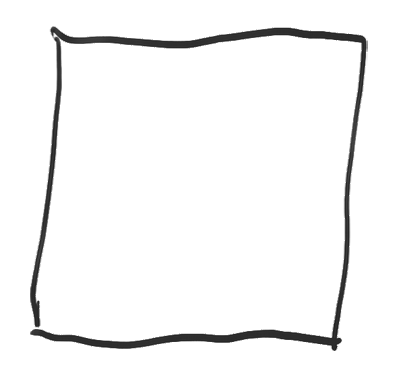}
    \caption{Rectangle}
  \end{subfigure}
  \begin{subfigure}[b]{0.2\linewidth}
    \includegraphics[width=\linewidth]{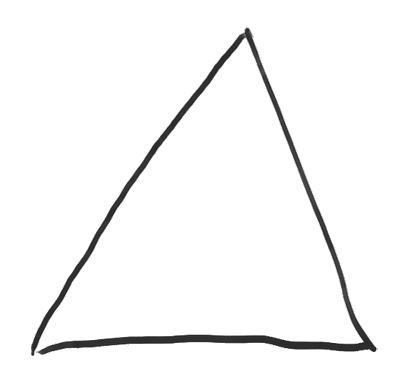}
    \caption{Triangle}
  \end{subfigure}
  \begin{subfigure}[b]{0.2\linewidth}
    \includegraphics[width=\linewidth]{cerad_rhombus.png}
    \caption{Diamond}
  \end{subfigure}
  \begin{subfigure}[b]{0.2\linewidth}
    \includegraphics[width=\linewidth]{cerad_rectangles.png}
    \caption{Rectangles}
  \end{subfigure}
  \begin{subfigure}[b]{0.2\linewidth}
    \includegraphics[width=\linewidth]{cerad_cube.png}
    \caption{Cube}
  \end{subfigure}  
  \begin{subfigure}[b]{0.2\linewidth}
    \includegraphics[width=\linewidth]{cerad_pentagrams.png}
    \caption{Pentagrams}
  \end{subfigure}  
  \caption{Set of gestures chosen from cognitive assessments.}
  \label{fig:assessment_symbols}
\end{figure}

Based on the design of sketching tasks in cognitive testing, we created a set of 11 gestures, which are commonly found in different cognitive assessments.
We focused on the geometric shapes of which tests are composed, e.g., the Clock Drawing Test contains a circle (clockface) and lines (hands).
The CERAD battery, MMSE and MoCA contain several shapes like pentagrams, diamonds and rectangles.
Single shapes in turn compose parts of other assessments, such as the ROCF depicted in figure \ref{fig:rocf_annotated}, which contains several sub-shapes, such as, triangles, rectangles, lines and circles.
As depicted in figure \ref{fig:assessment_symbols} a total of 8 shapes were chosen from the most commonly used cognitive assessments: arrow, circle, rectangle, triangle, circle, diamond, overlapping rectangles, cube and pentragrams. 
We chose 3 additional gestures based on a previously conducted user study, where we asked participants to specify gestures that they would use to indicate that they are finished with the current handwriting task.
Our symbol data set consists of 11 classes (shapes) with 100 samples per class.
The 7 subjects have provided a total of 7700 handwritten samples.

\begin{figure}
  \centering  
  \begin{subfigure}[b]{0.2\linewidth}
    \includegraphics[width=\linewidth]{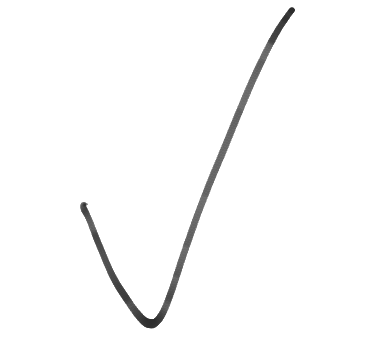}
    \caption{Checkmark}
  \end{subfigure}  
  \begin{subfigure}[b]{0.2\linewidth}
    \includegraphics[width=\linewidth]{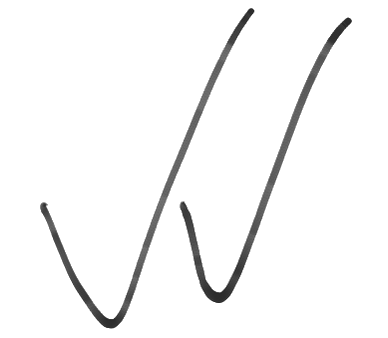}
    \caption{Checkmarks}
  \end{subfigure}  
  \begin{subfigure}[b]{0.2\linewidth}
    \includegraphics[width=\linewidth]{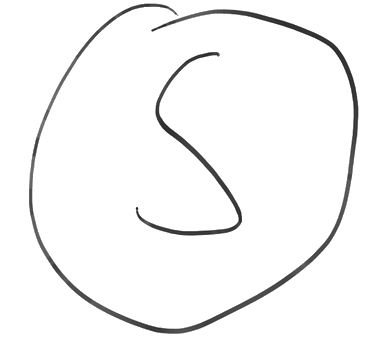}
    \caption{Send Symbol}
  \end{subfigure}  
  \caption{Set of symbols.}
  \label{fig:symbol_dataset}
\end{figure}

\subsection{Interakt Architecture}
In the Interakt use case the patient performs a digitalised cognitive assessment using a digital pen, which captures handwriting data in real-time. 
Figure \ref{fig:arch} shows the technical architecture, in which the digital ink is analysed using the previously described syntactic and semantic features.
Completing the cognitive assessment results in raw pen data being streamed to the backend service, where a document is created and indexed based on the performed test.
This document contains semantic information about the areas of the test (e.g., text fields, figures etc.) and the digital ink data.
We store the documents in a file format called \emph{XForm}, which is either a JSON or XML based structured description of the test and the captured ink.
With this format a visual representation of the completed test can be reconstructed and the doctor can retrace the patient's input using a playback functionality that replays the strokes in real-time as they were recorded.
Based on the respective assessment different sets of syntactic and semantic features are used by the \emph{pen data processing server} to analyse the handwritten and sketched contents of the test and deliver aggregated evaluation results that can be presented to the therapist.
Depending on the situation the analysis of the assessment may also involve additional patient data or previous test results, which are obtained from the \emph{data warehouse}.
The processed and evaluated assessment is finally also stored in the data warehouse, from where the doctor can access the results of the assessment in the \emph{therapist interface}.
The entire evaluation process takes place in real-time. 
\begin{figure}
\centering
\includegraphics[width=\textwidth]{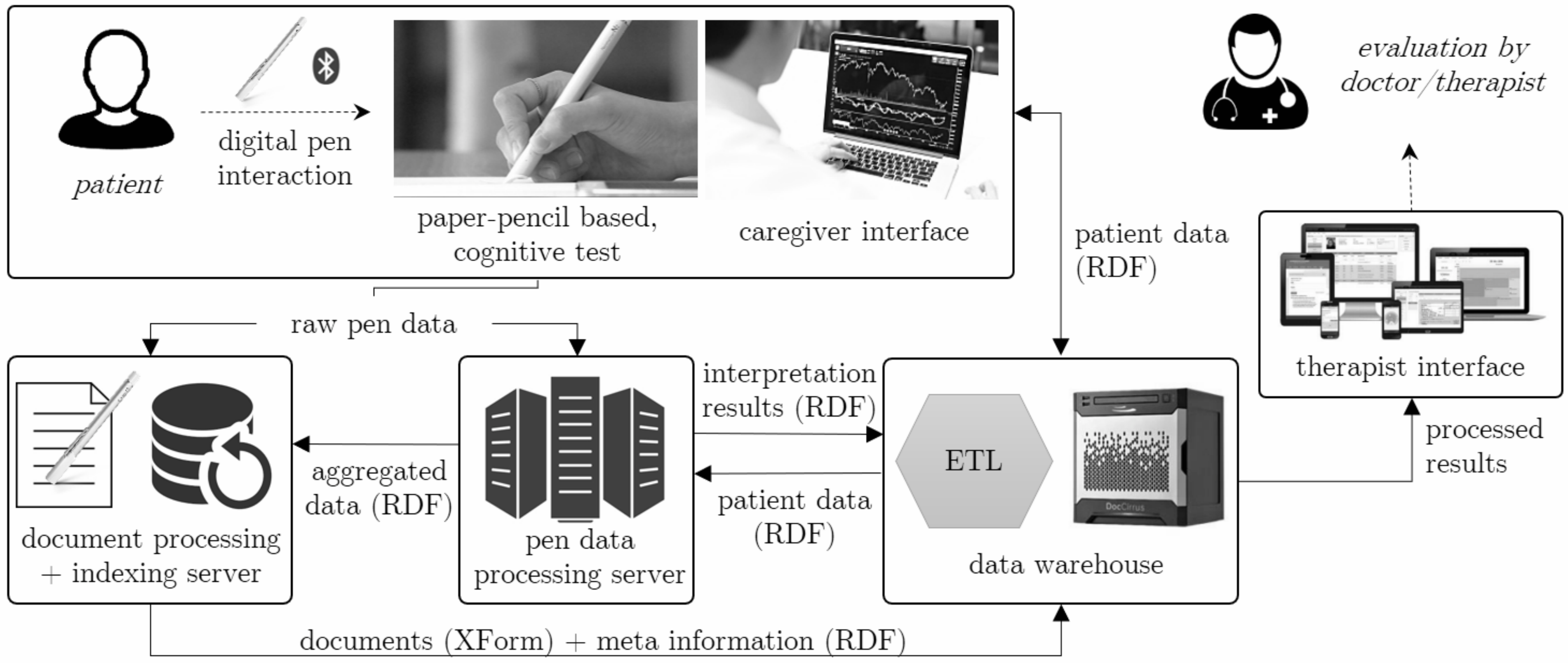}
  \caption{System Architecture in the dementia screening use case.}
  \label{arch}
\end{figure}

\section{Multimodality}

\begin{figure}
	\includegraphics[width=\linewidth,trim={6.1cm 1.2cm 6.5cm 1.2cm},clip]{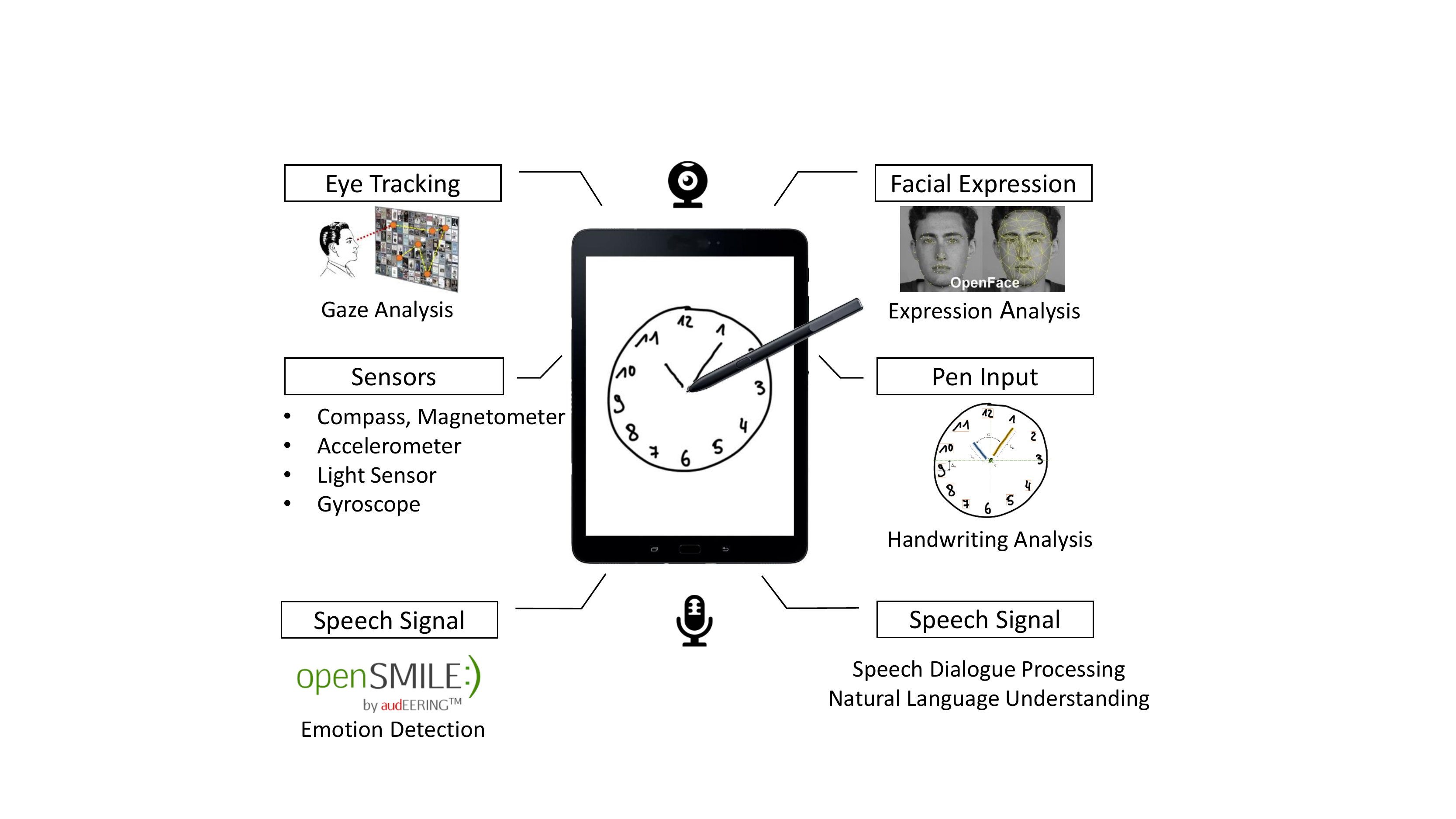}
	\caption{Multimodal interaction architecture on mobile device.}
	\label{fig:multimodal_tablet}
\end{figure}

In this section, we describe how additional modalities, beyond pen-based features, can help in the analysis of observed user behaviour, when interacting with a tablet computer and relying on the built-in sensors only.
For instance, researchers in the medical domain investigated ``observable differences in the communicative behaviour of patients with specific psychological disorders'' \cite{DeVault:2014:SKV:2617388.2617415}, e.g., the detection of depression from facial actions and vocal prosody \cite{Cohn2009}, which can be realised using the camera and microphone of a tablet device.
Including additional modalities can help with the disambiguation of signal- or semantic-level information in one error-prone recognition modality by using partial information supplied by another modality \cite{Oviatt2015}.
We consider the digital pen signal as primary modality for behaviour characterisation in combination with additional sensors and modalities as indicated in Figure \ref{fig:multimodal_tablet}: eye tracking and facial expression analysis via the video signal of the front-facing camera, natural speech processing via the built-in microphone and additional sensor inputs of modern tablet devices.



%

\subsection{Eye Tracking}
Eye tracking can improve human behaviour analysis, because human gaze is related to cognitive processes.
For instance, gaze trajectories can be used for inferring a user's task \cite{Yarbus1967}, for differentiating between novices and experts \cite{Bednarik:2012:EVA:2072690.2072965} and to model human visual attention \cite{Borji2013}.
Further, the number and duration of fixations and the transitions between different contents provide information about a user's cognitive engagement \cite{Lemaignan2016} and its cognitive load \cite{Korbach2017a}.
To augment pen signals, it is interesting that users pro-actively control their gaze behaviour to gather visual information for guiding movements across different activities \cite{Johansson2001} including hand movements \cite{Land1999}.
This relation suggests that pen and gaze signals can be analysed jointly for improving behaviour characterisation.
For multisensory behaviour analysis on unmodified mobile devices, RGB-based eye tracking is most interesting, because it's deployable using the built-in front-facing camera \cite{Wood2014}.
Compared to professional eye tracking equipment, the tracking quality is significantly lower \cite{Barz2017}. 
However, the form-factor can be essential for certain use cases, e.g., in dementia day hospitals that require non-obtrusive devices due to the patient's cognitive abilities.

\subsection{Facial Expressions}
OpenFace\footnote{https://github.com/TadasBaltrusaitis/OpenFace/} \cite{baltrusaitis_openface} is an open source toolkit for facial behaviour analysis using the stream of an RGB-webcam.
It provides state-of-the-art performance in facial landmark and head pose tracking, as well as facial action unit recognition which can be used to infer emotions.
These observations can be used for affective user interaction.
Further, it enables webcam-based eye tracking.

\subsection{Speech Signal}


The openSMILE toolkit\footnote{https://audeering.com/technology/opensmile/} \cite{Eyben:2013:RDO:2502081.2502224} provides methods for speech-based behaviour analysis and is distributed under an open source license. 
It offers an API for low-level feature extraction from audio signals and pre-trained classifiers for voice activity detection, speech-segment detection and speech-based emotion recognition in real-time.
The toolkit can be used on top of speech-based interaction frameworks to add a valence to users' utterances.


\section*{Acknowledgement}
This research is part of the Intera-KT project, which is supported by the German Federal Ministry of Education and Research (BMBF) under grant number 16SV7768.

\bibliographystyle{acm}
\bibliography{ink-features}

\newpage

\appendix
\section{Sonntag/Weber Features}
The features described in this section are implementations based on the 14 features described by Sonntag et al.~\cite{WEBER}. For this section we use the following notations:

A stroke is a sequence $S$ of samples,
\begin{equation}
    S = \left\{ \vec { s } _ { i } | i \in [ 0 , n - 1 ] , t _ { i } < t _ { i + 1 } \right\}
\end{equation}
where n is the number of recorded samples. A sequence of strokes is indicated by
\begin{equation}
    \mathcal{S} = \left\{ S _ { i } | i \in [ 0 , m - 1 ] \right\}
\end{equation}
where m is the number of strokes.

The centroid is defined as 
\begin{equation}
    \vec { \mu } = \frac { 1 } { n } \sum _ { i = 0 } ^ { n - 1 } \vec { S } _ { i }
\end{equation}
where n is the number of samples used for the classification, the mean radius (standard deviation) as
\begin{equation}
    \mu _ { r } = \frac { 1 } { n } \sum _ { i = 0 } ^ { n - 1 } \left\| \vec { s } _ { i } - \vec { \mu } \right\|
\end{equation}
and the angle as
\begin{equation}
    \varphi _ { s _ { i } } = \cos ^ { - 1 } \bigg\{ \frac { \left( s _ { i } - s _ { i - 1 } \right) \cdot \left( s _ { i + 1 } - s _ { i } \right) } { \left\| s _ { i } - s _ { i - 1 } \right\| \left\| s _ { i + 1 } - s _ { i } \right\| } \bigg\} 
\end{equation}

\subsection{Number of Strokes}
\begin{equation}
    f _ { 1 } = | \mathcal{ S } |
\end{equation}

\subsection{Length}
\begin{equation}
    f _ { 2 } = \sum _ { i = 0 } ^ { n - 2 } \left\| s _ { i } - s _ { i + 1 } \right\|
\end{equation}

\subsection{Area}
The area covered by the sequence of strokes $\mathcal{S}$ is defined as the area of the bounding box that results from a sequence of strokes.
We calculate the area of the convex hull $A$ based on Graham's algorithm\cite{graham1972}
\begin{equation}
    f _ { 3 } = Area ( Conv ( \mathcal{ S } ) ) = A
\end{equation}

\subsection{Perimeter Length}
The length of the path around the convex hull
\begin{equation}
    f _ { 4 } = \left\| Conv ( \mathcal{ S } ) \right\|
\end{equation}

\subsection{Compactness}
\begin{equation}
    f _ { 5 } = \frac { \left\| Conv ( \mathcal{ S } ) \right\| ^ { 2 } } { A }
\end{equation}

\subsection{Eccentricity}
Let $a$ and $b$ denote the length of the major or minor axis of the convex hull, respectively
\begin{equation}
    f _ { 6 } = \sqrt { 1 - \frac { b ^ { 2 } } { a ^ { 2 } } }
\end{equation}

\subsection{Principal Axes}
\begin{equation}
    f _ { 7 } = \frac { b } { a }
\end{equation}

\subsection{Circular Variance}
Let $\mu _ { r }$ denote the mean distance of the samples to the centroid $\mu$.
The circular variance is then computed as follows
\begin{equation}
    f _ { 8 } = \frac { 1 } { n \cdot {\mu _ { r }} ^ { 2 }  } \sum _ { i = 0 } ^ { n - 1 } \left( \left\|  s  _ { i } - \mu  \right\| - \mu _ { r } \right) ^ { 2 }
\end{equation}

\subsection{Rectangularity}
\begin{equation}
    f _ { 9 } = \frac { A } { a \cdot b }
\end{equation}

\subsection{Closure}
\begin{equation}
    f _ { 10 } = \frac { \left\|  s _ { 0 } -  s _ { n } \right\| } { f _ { 4 } }
\end{equation}

\subsection{Curvature}
Let $\varphi ( s _ { i } )$ be the angle between the $\overline { s _ { i - 1 } s _ { i } }$ and $\overline { s _ { i } s _ { i + 1 } }$ segments at $s_{i}$.
\begin{equation}
    f _ { 11 } = \sum _ { i = 1 } ^ { n - 2 } \varphi ( { s } _ { i } )
\end{equation}

\subsection{Perpendicularity}
\begin{equation}
    f _ { 12 } = \sum _ { i = 1 } ^ { n - 2 } \sin \left( \varphi ( s _ { i } ) \right) ^ { 2 }
\end{equation}

\subsection{Signed Perpendicularity}
\begin{equation}
    f _ { 13 } = \sum _ { i = 1 } ^ { n - 2 } \sin \left( \varphi ( s  _ { i } ) \right) ^ { 3 }
\end{equation}

\subsection{Angles after Equidistant Resampling}
For this feature we do an equidistant resampling with 6 line segments.
The five angles between succeeding lines are considered to make the features scale and rotation invariant (normalisation of writing speed).
\begin{equation}
    f _ { 14 } = \sum _ { i = 0 } ^ { 4 } \sin ( \alpha _ i ) , \sum _ { i = 0 } ^ { 4 } \cos ( \alpha )
\end{equation}

\section{Rubine's Features}
Features from this section are implementations of the described features by Rubine~\cite{RUBINE}.
\subsection{Cosine of initial angle}
\begin{equation}
    f _ { 1 } = \frac {( x _ { 2 } - x _ { 0 } )} { \sqrt { \left( x _ { 2 } - x _ { 0 } \right) ^ { 2 } + \left( y _ { 2 } - y _ { 0 } \right) ^ { 2 } } }
\end{equation}

\subsection{Sine of initial angle}
\begin{equation}
    f _ { 2 } = \frac {( y _ { 2 } - y _ { 0 } )}  {\sqrt { \left( x _ { 2 } - x _ { 0 } \right) ^ { 2 } + \left( y _ { 2 } - y _ { 0 } \right) ^ { 2 } } }
\end{equation}

\subsection{Length of bounding box diagonal}
\begin{equation}
    f _ { 3 } = \sqrt { \left( x _ { m a r } - x _ { m i n } \right) ^ { 2 } + \left( y _ { \max } - y _ { m i n } \right) ^ { 2 } }
\end{equation}

\subsection{Angle of the bounding box diagonal}
\begin{equation}
    f _ { 4 } = \arctan \frac { y _ { \max } - y _ { \min } } { x _ { \max } - x _ { \min } }
\end{equation}

\subsection{Distance between first and last point}
\begin{equation}
    f _ { 5 } = \sqrt { \left( x _ { n - 1 } - x _ { 0 } \right) ^ { 2 } + \left( y _ { n - 1 } - y _ { 0 } \right) ^ { 2 } }
\end{equation}

\subsection{Cosine of the angle between first and last point}
\begin{equation}
    f _ { 6 } = \frac {( x _ { n - 1 } - x _ { 0 } )} { f _ { 5 } }
\end{equation}

\subsection{Sine of the angle between first and last point}
\begin{equation}
    f _ { 7 } = \frac { ( y _ { P - 1 } - y _ { 0 } ) } { f _ { 5 } }
\end{equation}

\subsection{Total gesture length}
Let $\Delta x_{i} = x _ {i+1} - x _ {i}$, $\Delta y_{i} = y_{i+1} - y_{i}$
\begin{equation}
    f _ { 8 } = \sum _ { i = 0 } ^ { n - 2 } \sqrt { {\Delta x _ { i }} ^ { 2 } + {\Delta y _ { i }} ^ { 2 } }
\end{equation}

\subsection{Total angle traversed}
Let 
\begin{equation}
    \theta_{i} = arctan \frac{\Delta x_{i} \Delta y_{i-1} - \Delta x_{i-1} \Delta y_{i}}{\Delta x_{i} \Delta x_{i-1} + \Delta y_{i} \Delta y_{i-1}}
\end{equation}
\begin{equation}
    f _ { 9 } = \sum _ { i = 1 } ^ { n - 2 } \theta _ { i }
\end{equation}

\subsection{Sum of the absolute value of the angle at each point}
\begin{equation}
    f _ { 10 } = \sum _ { i = 1 } ^ { n - 2 } \left| \theta _ { i } \right|
\end{equation}

\subsection{Sum of the squared value of the angle at each point}
\begin{equation}
    f _ { 11 } = \sum _ { i = 1 } ^ { n - 2 } {\theta _ { i }} ^ { 2 }
\end{equation}

\subsection{Maximum speed (squared) of the gesture}
Let $\Delta t_{i} = t_{i+1} - t_{i}$
\begin{equation}
    f _ { 12 } = \max _ {i=0} ^ {n-2} \frac { {\Delta x_{i}} ^ {2} + {\Delta y_{i}} ^ {2}} { {\Delta t _ {i}} ^ {2} }
\end{equation}

\subsection{Duration of the gesture}
\begin{equation}
    f _ {13} = t _ {n - 1} - t_{0}
\end{equation}

\section{Features by Willems and Niels}
The features described in this section are implementations based on the feature set described by Willems and Niels~\cite{WILLEMS}. For this section we use the following notations:

Let $c$ be center of the bounding box around the gesture defined by the co-ordinate axes.
\begin{equation}
    c = 
        \left( 
            \begin{array} 
                { c } { x _ { c e n t e r } } \\ 
                      { y _ { c e n t e r } } 
            \end{array} 
        \right) 
        = 
        \left( 
            \begin{array} 
                { c } { x _ { m i n } + \frac { 1 } { 2 } \left( x _ { max } - x _ { min } \right) } \\ 
                      { y _ { m i n } + \frac { 1 } { 2 } \left( y _ { max } - y _ { min } \right) } 
            \end{array} 
        \right)
\end{equation}
While the ratio of the co-ordinate axis is not rotation independent, the ratio of the principal exes is. To determine the principal axes \emph{Principal Compnent Analysis} is used \cite{WILLEMS}.
Let $p_1$ and $p_2$ be the normalised principal component vectors of the set $S$. And let $c$ be the center of the box enclosing the trajectory and along the principal component vectors. The lengths of the major axes along the principal component vectors are
\begin{equation}
    \alpha = 2 \max _ { 0 \leq i < n} | p_2 \cdot (c - s_{i}) |, \quad \beta = 2 \max _ { 0 \leq i < n} | p_1 \cdot (c - s_{i}) |
\end{equation}

\subsection{Length of the gesture}
\begin{equation}
    f _ { 1 } = \sum _ { i = 0 } ^ { n - 2 } \left\| s _ { i + 1 } - s _ { i } \right\|
\end{equation}

\subsection{Area}
The area around the gesture is calculated using Graham's convex hull algorithm \cite{graham1972}.
\begin{equation}
    f _ { 2 } = A
\end{equation}

\subsection{Compactness}
Let $l$ be the the length of the perimeter of the convex hull, then compactness is defined as
\begin{equation}
    f _ { 3 } = \frac { l ^ { 2 } } { A }
\end{equation}

\subsection{Ratio between co-ordinate axes}
The lengths along the two co-ordinate axes (a, along the x-axis, and b along the y-axis) as given as
\begin{equation}
    a = \max _ { 0 \leq i < j < n } \left| x _ { i } - x _ { j } \right|
\end{equation}
\begin{equation}
    b = \max _ { 0 \leq i < j < n } \left| y _ { i } - y _ { j } \right|
\end{equation}
Eccentricity is a measure for the ratio between the co-ordinate axes.
\begin{equation}
    f _ { 4 } = \sqrt { 1 - \frac { b ^ { \prime 2 } } { a ^ { \prime 2 } } }
\end{equation}
where $a ^ { \prime } = a \wedge b ^ { \prime } = b \text { if } a > b \text { else } a ^ { \prime } = b \wedge b ^ { \prime } = a$.

\subsection{Ratio between co-ordinate axes}
The ratio of the co-ordinate axes, which is very much related to eccentricity, is denoted as follows
\begin{equation}
    f _ { 5 } = \frac { b ^ { \prime } } { a ^ { \prime } }
\end{equation}

\subsection{Closure}
\begin{equation}
    f _ { 6 } = \frac { f _ { 64 } } { f _ { 1 } } = \frac { \sum _ { i = 0 } ^ { n - 2 } \left\| s _ { i + 1 } - s _ { i } \right\| } { \left\| s _ { n - 1 } - s _ { 0 } \right\| }
\end{equation}

\subsection{Circular variance}
\begin{equation}
    f _ { 7 } = \frac { \sum _ { i = 0 } ^ { n - 1 } \left( | | s _ { i } - \mu | | - f _ { 68 } \right) ^ { 2 } } { n \cdot f _ { 68 } ^ { 2 } }
\end{equation}

\subsection{Curvature}
Let angle between sequenced samples be:
\begin{equation}
    \psi_{s_i} = \arccos\Bigg\{ 
    \frac{(s_i - s_{i-1}).(s_{i+1} - s_{i})}{\|s_i - s_{i-1} \| \|s_{i+1} - s_{i} \|}
    \Bigg\}
\end{equation}
then curvature will be:
\begin{equation} 
    f_{8} =  \sum_{i=1}^{n-2} {\psi_{s_i}}
\end{equation}

\subsection{Average curvature}
\begin{equation}
    f_{9} =  \frac{1}{n-2}\sum_{i=1}^{n-2} {\psi_{s_i}}
\end{equation}

\subsection{Standard deviation in curvature}
\begin{equation}
    f_{10} = \sqrt{
            \frac{1}{n-2}
            \sum_{i=1}^{n-2}
            { (\psi_{s_i} - f_{9} ) ^{2}} 
            }
\end{equation}

\subsection{Pen up/down ratio}
Let $\mathcal{G} = \{S_0, S_1, ..., S_{n-1}\}$ be the set of strokes composing a gesture of $n$ strokes.
The duration of stroke $S$ of length $m$ is given by
\begin{equation}
    \phi(S) = t_{m-1}(S) - t_{0}(S)
\end{equation}
where $t_j(S)$ is the $j$-th timestamp of stroke $S$.
The duration of all strokes $\mathcal{S}$ with $|\mathcal{S}|=n$ is then defined as
\begin{equation}
    \phi(\mathcal{S}) = \sum _ { i = 0 } ^ { n-1 } \phi(S_i)
\end{equation}
and the duration of the entire gesture is given by
\begin{equation}
    \phi(\mathcal{G}) = t_{|S_{n-1}|-1}(S_{n-1}) - t_0 (S_0)
\end{equation}
The ratio of pen up/down is the ratio between the time spent writing (pen down) and in air (pen up)
\begin{equation}
    f _ { 11 } = \frac{airtime}{writetime} = \frac{\phi(\mathcal{G}) - \phi(\mathcal{S})}{\phi(\mathcal{S})}          
\end{equation}

\subsection{Average direction}
\begin{equation}
    f_{12} =  \frac { 1 } { n-1 } \sum_{i=0}^{n-2} \arctan \frac { y_{i+1} -y_{i}} { x_{i+1} -x_{i} }
\end{equation}

\subsection{Perpendicularity}
\begin{equation}
    f_{13} = \sum_{i=1}^{n-2}
            \sin^2
            {\psi_{s_i}}
\end{equation}

\subsection{Average perpendicularity}
\begin{equation}
    f_{14} =  \frac{1}{n-2}
            \sum_{i=1}^{n-2}
            \sin^2
            {\psi_{s_i}}
\end{equation}

\subsection{Standard deviation in perpendicularity}
\begin{equation}
    f_{15} = \sqrt{
            \frac{1}{n-2}
            \sum_{i=1}^{n-2}
            { (\sin^2 {\psi_{s_i}} - f_{14} ) ^{2}} 
            }
\end{equation}

\subsection{Centroid offset}
The principal axes are used to calculate the centroid offset:
\begin{equation}
    f _ { 16 } = \left| \boldsymbol { p } _ { 2 } \cdot ( \boldsymbol { \mu } - \boldsymbol { c } ) \right|
\end{equation}

\subsection{Length of first principal axis}
Based on the principal axis, its length is another feature
\begin{equation}
    f _ { 17 } = \alpha
\end{equation}

\subsection{Sine orientation of principal axis}
The orientation of the principal axis $\psi$ is given by
\begin{equation}
    f _ { 18 } = \sin \psi = p _ { 1 _ { y } }
\end{equation}

\subsection{Cosine orientation of principal axis}
\begin{equation}
    f _ { 19 } = \cos \psi = p _ { 1 _ { x } }
\end{equation}

\subsection{Rectangularity}
Based on the lengths pf the major axes along the principal component vectors and the area of the convex hull $A$, the rectangularity is defined as:
\begin{equation}
    f _ { 20 } = \frac { A } { \alpha \cdot \beta }
\end{equation}

\subsection{Maximum angular difference}
\begin{equation}
    f_{21} = \max_{1+k \leq i \leq n-k} \psi_{s_i}^{k}
\end{equation}

\subsection{Average pressure}
\begin{equation}
    f _ { 22 } = \frac { 1 } { n } \sum _ { i = 0 } ^ { n - 1 } p _ { i }
\end{equation}

\subsection{Standard deviation of pressure}
\begin{equation}
    f _ { 23 } = \sqrt { \frac { 1 } { n } \sum _ { i = 0 } ^ { n - 1 } \left( p _ { i } - f _ { 22 } \right) ^ { 2 } }
\end{equation}

\subsection{Duration}
\begin{equation}
    f _ { 24 } = t _ { n - 1 } - t _ { 0 }
\end{equation}

\subsection{Average velocity}
\begin{equation}
    \boldsymbol { v } _ { i } = \frac { \left\| s _ { i + 1 } - s _ {i} \right\| + \left\| s _ {i} - s _ { i - 1 } \right\| } { t _ { i + 1 } - t _ { i - 1 } }
\end{equation}

\begin{equation}
    f _ { 25 } = \frac { 1 } { n - 2 } \sum _ { i = 1 } ^ { n - 2 } \left\| \boldsymbol { v } _ { i } \right\|
\end{equation}

\subsection{Standard deviation of velocity}
\begin{equation}
    f _ { 26 } = \sqrt { 
        \frac { 1 } { n - 2 } \sum _ { i = 1 } ^ { n - 2 } \left( \left\| \boldsymbol { v } _ { i } \right\| - f _ { 25 } \right) ^ { 2 } 
    }
\end{equation}

\subsection{Maximum velocity}
\begin{equation}
    f _ { 27 } = \max _ { 1 \leq i \leq n - 2 } | | \boldsymbol { v } _ { i } | |
\end{equation}

\subsection{Average acceleration}
\begin{equation}
    \boldsymbol { a } _ { i } = 
        \frac { 
            \boldsymbol { v } _ { i + 1 } - \boldsymbol { v } _ { i - 1 } 
        } { 
            t _ { i + 1 } - t _ { i - 1 } 
        }
\end{equation}

\begin{equation} 
    f _ { 28 } = \frac { 1 } { n - 4 } \sum _ { i = 2 } ^ { n - 3 } \left\| \boldsymbol { a } _ { i } \right\|
\end{equation}

\subsection{Standard deviation of acceleration}
\begin{equation}
    f _ { 29 } = \sqrt { \frac { 1 } { n - 4 } \sum _ { i = 2 } ^ { n - 3 } ( \| \boldsymbol { a } _ { i } | | - f _ { 28 } ) ^ { 2 } }
\end{equation}

\subsection{Maximum acceleration}
\begin{equation}
    f _ { 30 } = \max _ { 2 \leq i \leq n - 3 } \left\| \boldsymbol { a } _ { i } \right\|
\end{equation}

\subsection{Minimum acceleration}
\begin{equation}
    f _ { 31 } = \min _ { 2 \leq i \leq n - 3 } \left\| \boldsymbol { a } _ { i } \right\|
\end{equation}

\subsection{Number of cups}

\begin{equation}
    f _ { 32 } = nCups
\end{equation}

\subsection{Offset of the first cup}

\begin{equation}
    f _ { 33 } = lastCupOffset
\end{equation}

\subsection{Offset of the last cup}

\begin{equation}
    f _ { 34 } = firstCupOffset
\end{equation}

\subsection{Initial horizontal offset}
\begin{equation}
    f _ { 35 } = \frac { x _ { 0 } - x _ { \min } } { a }
\end{equation}

\subsection{Final horizontal offset}
\begin{equation}
    f _ { 36 } = \frac { x _ { n - 1 } - x _ { \min } } { a }
\end{equation}

\subsection{Initial vertical offset}
\begin{equation}
    f _ { 37 } = \frac { y _ { 0 } - y _ { \min } } { b }
\end{equation}

\subsection{Final vertical offset}
\begin{equation}
    f _ { 38 } = \frac { y _ { n - 1 } - y _ { \min } } { b }
\end{equation}

\subsection{Number of straight lines}
Based on the definition of straight lines by Willems and Niels\cite{WILLEMS}, we denote the set of straight lines $L_i$ inside a gesture as $\mathcal{L} = \{L_0, L_1, ..., L_{i-1}\}$ and the number of straight lines as

\begin{equation}
    f _ { 39 } = | \mathcal { L } |
\end{equation}

\subsection{Average length of straight lines}
Let $\left\| L _ {i} \right\|$ be the length of a straight line, then the average length of straight lines is calculated as
\begin{equation}
    f _ { 40 } = \frac { 1 } { | \mathcal { L } | } \sum _ { i = 0 } ^ { | \mathcal { L } | - 1 } \left\| L _ { i } \right\|
\end{equation}

\subsection{Standard deviation of straight line length}
\begin{equation}
    f _ { 41 } = \sqrt { 
        \frac { 1 } { | \mathcal { L } |  } \sum _ { i = 0 } ^ { | \mathcal { L } | - 1 } \left( \left\| L _ { i } \right\| - f _ { 40 } \right) ^ { 2 } 
    }
\end{equation}

\subsection{Straight line ratio}
\begin{equation}
    f _ { 42 } = \sum _ { i = 0 } ^ { | \mathcal { L } | - 1} \frac { \left\| L _ {i} \right\| } { \sum _ { j = 1 } ^ { n - 1 } \left\| s _ { j } - s _ { j - 1 } \right\| } 
        = \frac { 1 } { f _ { 1 } } \sum _ { i = 0 } ^ { | \mathcal { L } | - 1 } \left\| L _ { i } \right\| 
\end{equation}

\subsection{Largest straight line ratio}
\begin{equation}
    f _ { 43 } = \max _ { 0 \leq i \leq n - 1 } \frac { \left\| L _ { i } \right\| } { \sum _ { j = 1 } ^ { n - 1 } \left\| s _ { j } - s _ { j - 1 } \right\| } = \frac { 1 } { f _ { 1 } } \max _ { 0 \leq i < n < | \mathcal { L } | } \left\| L _ { i } \right\|
\end{equation}

\subsection{Number of pen down events}
Let $\mathcal{G} = \{S_0, S_1, ..., S_{i-1}\}$ be the set of strokes composing a gesture of $n$ strokes. 
The number of pen down events equals the number of strokes
\begin{equation}
    f _ { 44 } = | \mathcal{G} |
\end{equation}

\subsection{Octants}
\begin{equation} 
  f_{44+o} =    \frac{1}{n-1}
                \sum_{i=0}^{n-1}
                \omega_{io}
\end{equation}

where
\begin{equation} 
\omega_{io}=\begin{cases}
            1 & \text{if $\frac{\pi}{4} (o-1) \leq \nu_i < \frac{\pi}{4} o$ }\\
            0 & \text{if $\nu_i <  \frac{\pi}{4} (o-1) \lor \nu_i \geq \frac{\pi}{4} o$}
         \end{cases}
\end{equation}

and where
\begin{equation} 
\nu_i= \arctan \dv{y_i}{x_i}
\end{equation}

and
\begin{align*}
dx_i= x_i - x_{center} \\
dy_i= y_i - y_{center}    
\end{align*}

\subsection{Number of connecting strokes}
We define the set of connected components as $\mathcal{C} = \{C_0, C_1, ..., C_{i-1}\}$.
A connected component $C_i$ is a part of a gesture that consists of one or more strokes that touch each other, and that do not touch any other strokes \cite{WILLEMS}. 
\begin{equation}
    f _ { 53 } = | \mathcal{C} |
\end{equation}

\subsection{Number of crossings}
\begin{equation} 
  f_{54} =  \sum_{i=1}^{n-2}
            \sum_{j=i+1}^{n-1}
            \kappa_{ij}
\end{equation}
where
\begin{equation} 
\kappa_{ij}=\begin{cases}
            1 & \text{if $s_i \rightarrow s_{i+1} \cap s_j \rightarrow s_{j+1} \neq \emptyset $ }\\
            0 & \text{if $s_i \rightarrow s_{i+1} \cap s_j \rightarrow s_{j+1} = \emptyset $}
         \end{cases}
\end{equation}

\subsection{Cosine of initial angle}
\begin{equation}
    f_{55} =  \frac {  x _ { 2 } - x _ { 0 } } { \|s _ { 2 } - s _ { 0 } \|}  
\end{equation}

\subsection{Sine of initial angle}
\begin{equation}
    f_{56} =  \frac { y _ { 2 } - y _ { 0 }  } { \|s _ { 2 } - s _ { 0 } \|}  
\end{equation}

\subsection{Lenght of the bounding box diagonal}
Given the two co-ordinate axes the length of the bounding box is given as
\begin{equation}
    f _ { 57 } = \sqrt { a ^ { 2 } + b ^ { 2 } }
\end{equation}

\subsection{Angle of the bounding box diagonal}
\begin{equation}
    f_{58} =   \tan  \frac { b } { a }
\end{equation}

\subsection{Length between first and last point}
\begin{equation}
    f _ { 59 } = \left\| s _ { n-1 } - s _ { 0 } \right\|
\end{equation}

\subsection{Cosine of first to last point}
\begin{equation}
    f_{60} =  \frac {  x _ { n-1 } - x _ { 0 } } { \|s _ { n-1 } - s _ { 0 } \|}  
\end{equation}

\subsection{Sine of first to last point}
\begin{equation}
    f_{61} =  \frac {  y _ { n-1 } - y _ { 0 } } { \|s _ { n-1 } - s _ { 0 } \|}  
\end{equation}

\subsection{Absolute curvature}
\begin{equation}
    f_{62} = \sum_{i=1}^{n-2}
            {| \psi_{s_i}|}
\end{equation}

\subsection{Squared curvature}
\begin{equation}
    f_{63} = \sum_{i=1}^{n-2}
            {\psi_{s_i}^{2}}
\end{equation}

\subsection{Macro perpendicularity}
Let angle between sampled points be:
\begin{equation} 
  \psi_{s_i}^{k} = \arccos\Bigg\{ 
  \frac{(s_i - s_{i-k}).(s_{i+k} - s_{i})}{\|s_i - s_{i-k} \| \|s_{i+k} - s_{i} \|}
  \Bigg\}
\end{equation}
then macro perpendicularity  will be:
\begin{equation} 
  f_{64} =  \sum_{i=1+k}^{n-k}
                \sin^2 
                {\psi_{s_i}^{k}}
\end{equation}

\subsection{Average macro perpendicularity}
\begin{equation} 
  f_{65} =      \frac{1}{n-2k}
                \sum_{i=1+k}^{n-k}
                \sin^2 
                {\psi_{s_i}^{k}}
\end{equation}

\subsection{Standard deviation in macro perpendicularity}
\begin{equation} 
  f_{66} = \sqrt{
            \frac{1}{n-2k}
            \sum_{i=1+k}^{n-k}
            { (\sin^2 {\psi_{s_i}^{k}} - f_{69} ) ^{2}} 
            }
\end{equation}

\subsection{Ratio of principal axes}
Based on the lengths pf the major axes along the principal component vectors, the ratio of the principal axes becomes:
\begin{equation}
    f _ { 67 } = \frac { \beta } { \alpha }
\end{equation}

\subsection{Average centroidal radius}
The average distance of sample points from the centroid is a feature called average centroidal radius.
\begin{equation}
    f _ { 68 } = \frac { 1 } { n } \sum _ { i = 0 } ^ { n - 1 } \left\| s _ { i } - \mu \right\|
\end{equation}

\subsection{69 Standard deviation of the centroidal radius}
\begin{equation}
    f _ { 69 } = \sqrt { \frac { 1 } { n } \sum _ { i = 0} ^ { n - 1 } \left( \left\| s _ { i } - \mu \right\| - f _ { 68 } \right) ^ { 2 } }
\end{equation}

\subsection{Chain codes}
Let the chain code be defined as
\begin{equation} 
C_{s}=\begin{cases}
            1 & \text{if $ 0 \leq \psi_s < \frac{\pi}{4}  $ }\\
            2 & \text{if $ \frac{\pi}{4} \leq \psi_s < \frac{\pi}{2}  $ }\\
            3 & \text{if $ \frac{\pi}{2} \leq \psi_s < \frac{3\pi}{4}  $ }\\
            4 & \text{if $ \frac{3\pi}{4} \leq \psi_s < \pi  $ }\\
            5 & \text{if $ \pi \leq \psi_s < \frac{5\pi}{4}  $ }\\
            6 & \text{if $ \frac{5\pi}{4} \leq \psi_s < \frac{3\pi}{2}  $ }\\
            7 & \text{if $ \frac{3\pi}{2} \leq \psi_s < \frac{7\pi}{4}  $ }\\
            8 & \text{if $ \frac{7\pi}{4} \leq \psi_s < 2\pi  $ }\\
         \end{cases}
\end{equation}
then the average angle of the chain code will be
$$
\psi_{C_s} = \frac{(C_s - \frac{1}{2})\pi}{4}
$$
then
$$
f_{68+2s} = \sin{\psi_{C_s}}
$$
and
$$
f_{69+2s} = \cos{\psi_{C_s}}
$$

\subsection{Average stroke length}
If $S _ i \in \mathcal{S}$ is a stroke with $n$ sample points, then let $L _ i$ be the length of that stroke:
\begin{equation}
    L _ { i } = \sum _ { j = 0 } ^ { n - 2 } \left\| s _ { j + 1 } - s _ { j } \right\|
\end{equation}
Assuming $| \mathcal{S} | = m$ the average stroke length is given by
\begin{equation}
    f _ { 86 } = \frac {1} {m} \sum _ { i = 0 } ^ { m - 1 } L _ { i }
\end{equation}

\subsection{Standard deviation in stroke length}
\begin{equation}
    f _ { 87 } = \sqrt { \frac {1} {m} \sum _ { i = 0 } ^ { m - 1 } ( L _ { i } - f _ { 86 }) ^ 2 }
\end{equation}

\subsection{Average stroke direction}
If $S _ i \in \mathcal{S}$ is a stroke with $n$ sample points, then let $\varphi _ i$ be the direction of that stroke:
\begin{equation}
    \varphi _ i = \frac { 1 } { n-1 } \sum_{j=0}^{n-2} \arctan \frac { y_{j+1} -y_{j}} { x_{j+1} -x_{j} }
\end{equation}
Assuming $| \mathcal{S} | = m$ the average stroke direction is given by
\begin{equation}
    f_{88} = \frac{1}{m} \sum _ {i=0} ^ {m - 1} \varphi _ {i}
\end{equation}

\subsection{Standard deviation in stroke direction}
\begin{equation}
    f _ {89} = \sqrt{ \frac{1}{m} \sum _ {i=0} ^ {m - 1} ( \varphi _ {i} - f_{88} ) ^ 2 }
\end{equation}

\section{HBF49 Features}
The features described in this section are implementations based on the feature set HBF49 described by Delaye and Anquetil~\cite{HBF49}. For this section we use the following notations:

Let $B$ be the rectangular bounding box defined by $x_{min}, x_{max}, y_{min}, y_{max}$. The width $w$ and height $h$ of this box are defined as
\begin{equation}
    w = x_{max} - x_{min}, h = y_{max} - y_{min}
\end{equation}
Coordinates $c_x$ and $c_y$ are the coordinates of the center point $c$ of bounding box $B$.

Let $L_{i,j}$ be the length of the path between sample points $s_{i}$ and $s_{j}$, then $L$ is the total length of the path of the gesture.

Let $\mathcal{S}$ be the set of strokes composing the gesture.

\subsection{Horizontal position of first point}
Let $l = m a x ( h , w )$ be the side of a square box centered on $c$. The normalised position of the first point is then given by
\begin{equation}
f _ { 1 } = \frac { x _ { 0 } - c _ { x } } { l } + \frac { 1 } { 2 } 
\end{equation}

\subsection{Vertical position of first point}
\begin{equation}
    f _ { 2 } = \frac { y _ { 0 } - c _ { y } } { l } + \frac { 1 } { 2 }
\end{equation}

\subsection{Horizontal position of last point}
\begin{equation}
    f _ { 3 } = \frac { x _ { n - 1 } - c _ { x } } { l } + \frac { 1 } { 2 }  
\end{equation}

\subsection{Vertical position of last point}
\begin{equation}
    f _ { 4 } = \frac { y _ { n - 1 } - c _ { y } } { l } + \frac { 1 } { 2 }
\end{equation}

\subsection{First point to last point vector length}
\begin{equation}
    v = \overrightarrow{ s _ {0} s _ {n-1}}
\end{equation}
\begin{equation}
f _ { 5 } = \| v \|
\end{equation}

\subsection{Sine of first point to last point vector}
Let $u_x$ be the unit vector codirectional with the x axis.
\begin{equation}
    f _ { 6 } = \frac { { v } _ { x } \cdot { u _ { x } } } { f _ { 5 } }
\end{equation}

\subsection{Cosine of first point to last point vector}
Let $u_y$ be the unit vector codirectional with the y axis.
\begin{equation}
    f _ { 7 } = \frac {  { v _ { y } } \cdot  { u _ { y } } } { f _ { 5 } }
\end{equation}

\subsection{Closure}
\begin{equation}
    f _ { 8 } = \frac { \| v \| } { L }.
\end{equation}

\subsection{Sine of initial angle}
Our initial vector between the first and third point is given by $w = \overrightarrow { s _ { 0 } s _ { 2 } }$.
\begin{equation}
    f _ { 9 } = \frac { { w _ { x } } \cdot { u _ { x } } } { \| w \| } 
\end{equation}

\subsection{Cosine of initial angle}
\begin{equation}
    f _ { 10 } = \frac { { w _ { y } } \cdot { u _ { y } } } { \| w \| }
\end{equation}

\subsection{Horizontal inflexion}
Let $s_{m}$ be the middle-path point with respect to the middle point of segment $s _ { 0 } s _ { n-1 }$.
\begin{equation}
    f _ { 11 } = \frac { 1 } { w } \left( x _ { m } - \frac { x _ { 1 } + x _ { n } } { 2 } \right) 
\end{equation}

\subsection{Vertical inflexion}
\begin{equation}
    f _ { 12 } = \frac { 1 } { h } \left( y _ { m } - \frac { y _ { 1 } + y _ { n } } { 2 } \right)
\end{equation}

\subsection{Downstroke proportion}
Downstrokes are portions of drawing trajectories oriented towards the bottom of the writing surface, i.e. oriented towards increasing values in dimension y. 
\begin{equation}
    f _ { 13 } = \sum _ { k = 0 } L _ { k } \quad \{\forall \text{ } k \in \mathcal{S} \text{ } | \text{ } k \text{ is a downstroke}\}
\end{equation}

\subsection{Number of strokes}
\begin{equation}
f _ { 14 } = | \mathcal{S} |
\end{equation}

\subsection{Angle of the bounding box diagonal}
\begin{equation}
    f _ { 15 } = \arctan \frac { h } { w }
\end{equation}

\subsection{Trajectory length}
\begin{equation}
    f _ { 16 } = L
\end{equation}

\subsection{Ratio between bounding box and trajectory length}
\begin{equation}
    f _ { 17 } = \frac { w + h } { L }
\end{equation}

\subsection{Deviation}
\begin{equation}
    f _ { 18 } = \frac { 1 } { n } \sum _ { i = 0 } ^ { n - 1 } \left\| s _ { i } \mu \right\|
\end{equation}

\subsection{Average direction}
\begin{equation}
    f _ { 19 } = \frac { 1 } { n - 1 } \sum _ { i = 0 } ^ { n - 2 } \arctan \left( \frac { y _ { i + 1 } - y _ { i } } { x _ { i + 1 } - x _ { i } } \right)
\end{equation}

\subsection{Curvature}
\begin{equation}
    \theta _ { i } = \arccos
        \frac 
            { \overrightarrow{ s _ { i - 1 } s _ { i } } \cdot \overrightarrow { s _ { i } s _ { i + 1 } } } 
            { \| \overrightarrow{s _ { i - 1 } s _ { i }} \| \|  \overrightarrow{s _ { i } s _ { i + 1 }} \| }
\end{equation}
\begin{equation}  
    f _ { 20 } = \sum _ { i = 1 } ^ { n - 2 } \theta _ { i }
\end{equation}

\subsection{Perpendicularity}
\begin{equation}
    f _ { 21 } = \sum _ { i = 1 } ^ { n - 2 } \sin ^ { 2 } \left( \theta _ { i } \right)
\end{equation}

\subsection{k-Perpendicularity}
\begin{equation}
    \theta _ { i } ^ k = \arccos
        \frac 
            { \overrightarrow{ s _ { i - k } s _ { i } } \cdot \overrightarrow { s _ { i } s _ { i + k } } } 
            { \| \overrightarrow{s _ { i - k } s _ { i }} \| \|  \overrightarrow{s _ { i } s _ { i + k }} \| }
\end{equation}
\begin{equation}
    f _ { 22 } = \sum _ { i = k } ^ { n - k - 1 } \sin ^ { 2 } \left( \theta _ { i } ^ k \right)
\end{equation}

\subsection{k-Angle}
\begin{equation}
    f _ { 23 } = \max _ {i = k} ^ { n - k - 1 } \theta _ { i } ^ { k }
\end{equation}

\subsection{Dominant direction}
Let $n_a$ be the number of segments in $\mathcal{S}$ ($n_a = n - K$, with $K$ the number of strokes).
\begin{equation}
    f _ { 24 } = \frac { h _ { 1 } + h _ { 5 } } { n _ { a } }
\end{equation}
\begin{equation}
    f _ { 25 } = \frac { h _ { 2 } + h _ { 6 } } { n _ { a } }
\end{equation}
\begin{equation}
    f _ { 26 } = \frac { h _ { 3 } + h _ { 7 } } { n _ { a } }
\end{equation}
\begin{equation}
    f _ { 27 } = \frac { h _ { 4 } + h _ { 8 } } { n _ { a } }
\end{equation}

\subsection{Local changes in direction}
Local angle benefit from smoothing by linear combination of $\theta _ { i }$ and $\theta _ { i } ^ { k }$ (refer to $f_{14}$ and $f_{16}$):
\begin{equation}
    \psi _ { i } ^ { k } = \gamma \theta _ { i } + ( 1 - \gamma ) \theta _ { i } ^ { k }
\end{equation}
In the HBF49 feature set \cite{HBF49} the values are set empirically to $\gamma = 0.25$ and $k = 2$. The contributions of $\psi _ { i } ^ { k }$ angles are accumulated in four histogram bins uniformly distributed in $[0, \pi]$. Contributions to the histogram are weighted by the inverse of their angular distance with the central direction of the two neighboring bins. The features are obtained from the histogram $h$ divided by $n_a$ (see $f_{24}$-$f_{27}$).
\begin{equation}
    f _ { 28 } = \frac {h[0]} {n_a}
\end{equation}
\begin{equation}
    f _ { 29 } = \frac {h[1]} {n_a}
\end{equation}
\begin{equation}
    f _ { 30 } = \frac {h[2]} {n_a}
\end{equation}
\begin{equation}
    f _ { 31 } = \frac {h[3]} {n_a}
\end{equation}

\subsection{2D Histogram}
For the 2D histogram we devide the rectangular bounding box around the figure into $3 \times 3$ partitions of equal size.
Sampling points are sorted into the 9 cells resulting from partitioning \cite{2d_zoning}.
For each sample point a fuzzy weighted contribution to the 4 neigboring cells is computed, where the weights depend on the distance from the point to the cell centers \cite{nonrigid_feature_extraction}.

\begin{equation}
    f _ { 32 } = \frac { 1 } { n } \sum _ { i = 0 } ^ { n - 1 } \mu _ { 11 } ( s _ { i } )
\end{equation}
\begin{equation}
    f _ { 33 } = \frac { 1 } { n } \sum _ { i = 0 } ^ { n - 1 } \mu _ { 12 } ( s _ { i } )
\end{equation}
\begin{equation}
    f _ { 34 } = \frac { 1 } { n } \sum _ { i = 0 } ^ { n - 1 } \mu _ { 13 } ( s _ { i } )
\end{equation}
\begin{equation}
    f _ { 35 } = \frac { 1 } { n } \sum _ { i = 0 } ^ { n - 1 } \mu _ { 21} ( s _ { i } )
\end{equation}
\begin{equation}
    f _ { 36 } = \frac { 1 } { n } \sum _ { i = 0 } ^ { n - 1 } \mu _ { 22 } ( s _ { i } )
\end{equation}
\begin{equation}
    f _ { 37 } = \frac { 1 } { n } \sum _ { i = 0 } ^ { n - 1 } \mu _ { 23 } ( s _ { i } )
\end{equation}
\begin{equation}
    f _ { 38 } = \frac { 1 } { n } \sum _ { i = 0 } ^ { n - 1 } \mu _ { 31 } ( s _ { i } )
\end{equation}
\begin{equation}
    f _ { 39 } = \frac { 1 } { n } \sum _ { i = 0 } ^ { n - 1 } \mu _ { 32 } ( s _ { i } )
\end{equation}
\begin{equation}
    f _ { 40 } = \frac { 1 } { n } \sum _ { i = 0 } ^ { n - 1 } \mu _ { 33 } ( s _ { i } )
\end{equation}

\subsection{Hu moments}
Let $\mu = \left( \mu _ { x } , \mu _ { y } \right)$ be the center of gravity, then the central inertia moments are computed as follows
\begin{equation}
    m _ { p q } = \sum _ { i = 1 } ^ { n } \left( x _ { i } - \mu _ { x } \right) ^ { p } \left( y _ { i } - \mu _ { y } \right) ^ { q } , \quad \text { for } 0 \leq p , q \leq 3
\end{equation}
In order to guarantee scale independence the moments are normalised:
\begin{equation}
    \nu _ { p q } = \frac { m _ { p q } } { m _ { 00 } ^ { \gamma } } , \quad \text { with } \gamma = 1 + \frac { p + q } { 2 }
\end{equation}
The seven Hu moments \cite{hu_moments} are computed as:
\begin{equation}
    f _ { 41 } = \nu _ { 02 } + \nu _ { 20 }
\end{equation}
\begin{equation}
    f _ { 42 } = \left( \nu _ { 20 } - \nu _ { 02 } \right) ^ { 2 } + 4 \nu _ { 11 } ^ { 2 }
\end{equation}
\begin{equation}
    f _ { 43 } = \left( \nu _ { 30 } - 3 \nu _ { 12 } \right) ^ { 2 } + \left( 3 \nu _ { 21 } - \nu _ { 03 } \right) ^ { 2 }
\end{equation}
\begin{equation}
    f _ { 44 } = \left( \nu _ { 30 } + \nu _ { 12 } \right) ^ { 2 } + \left( \nu _ { 21 } + \nu _ { 03 } \right) ^ { 2 }
\end{equation}
\begin{equation}
    \left. 
    \begin{aligned}  
        f _ { 45 } = & \left( \nu _ { 30 } - 3 \nu _ { 12 } \right) ^ { 2 } \left( \nu _ { 30 } + \nu _ { 03 } \right)    \left[ \left( \nu _ { 30 } + \nu _ { 12 } \right) ^ { 2 } - 3 \left( \nu _ { 21 } + \nu _ { 03 } \right) ^ { 2 } \right]  \\ 
        &  + \left( 3 \nu _ { 21 } - \nu _ { 03 } \right) \left( \nu _ { 21 } + \nu _ { 03 } \right)  \left[ 3 \left( \nu _ { 30 } + \nu _ { 12 } \right) ^ { 2 } - \left( \nu _ { 21 } + \nu _ { 03 } \right) ^ { 2 } \right] 
    \end{aligned} 
    \right.
\end{equation}
\begin{equation}
    \left.
    \begin{aligned} 
        f _ { 46 } = & \left( \nu _ { 20 } - \nu _ { 02 } \right) \left[ \left( \nu _ { 30 } + \nu _ { 12 } \right) ^ { 2 } - \left( \nu _ { 21 } + \nu _ { 03 } \right) ^ { 2 } \right] \\ & + 4 \nu _ { 11 } \left( \nu _ { 30 } + \nu _ { 12 } \right) \left( \nu _ { 21 } + \nu _ { 03 } \right) 
    \end{aligned} 
    \right.
\end{equation}
\begin{equation}
    \left.
    \begin{aligned} 
        f _ { 47 } = & \left( 3 \nu _ { 21 } - \nu _ { 03 } \right) \left( \nu _ { 30 } + \nu _ { 12 } \right) \left[ \left( \nu _ { 30 } + \nu _ { 12 } \right) ^ { 2 } - 3 \left( \nu _ { 21 } + \nu _ { 03 } \right) ^ { 2 } \right] \\ & - \left( \nu _ { 30 } - 3 \nu _ { 12 } \right) \left( \nu _ { 21 } + \nu _ { 03 } \right)  \left[ 3 \left( \nu _ { 30 } + \nu _ { 12 } \right) ^ { 2 } - \left( \nu _ { 21 } + \nu _ { 03 } \right) ^ { 2 } \right] \\
    \end{aligned}
    \right.
\end{equation}

\subsection{Normalised convex hull area}
Let $H$ be the convex hull around the gesture, then $A_H$ denotes the area of the convex hull.
\begin{equation}
    f _ { 48 } = \frac { A _ { H } } { w * h }
\end{equation}

\subsection{Compactness}
\begin{equation}
    f _ { 49 } = \frac { L ^ { 2 } } { A _ { H } }
\end{equation}

\end{document}